\documentclass[letterpaper, 10 pt, conference]{ieeeconf}  % Comment this line out if you need a4paper

\IEEEoverridecommandlockouts                              % This command is only needed if 
\usepackage{xcolor}
\usepackage{graphicx}
\usepackage{caption}
\usepackage{subcaption}
\usepackage{bm}
\usepackage{multirow}
\usepackage{amssymb}
\usepackage[utf8]{inputenc}
\usepackage[english]{babel}
\usepackage{amsthm}
\usepackage{amsmath}
\usepackage[normalem]{ulem}
\usepackage{cases}

\newtheorem{theorem}{Theorem}[section]

\newtheorem{lemma}[theorem]{Lemma}

\title{\LARGE \bf
SwarmCCO: Probabilistic Reactive Collision Avoidance for Quadrotor Swarms under Uncertainty
}

\author{Senthil Hariharan Arul$^{1}$ and Dinesh Manocha$^{2}$% <-this % stops a space
\thanks{*This work was supported in part by ARO under Grants W911NF1810313 and W911NF1910315, and in part by Intel.}% <-this % stops a space
\thanks{$^{1}$Senthil Hariharan Arul is with the Department of Electrical and Computer Engineering,
        University of Maryland at College Park, MD, USA
        {\tt\small sarul1@umd.edu}}%
\thanks{$^{2}$Dinesh Manocha is with the Departments of Computer Science and Electrical \& Computer Engineering, University of Maryland at College Park, MD, USA
        {\tt\small dm@cs.umd.edu}}%
}

\begin{document}

\maketitle
\thispagestyle{empty}
\pagestyle{empty}

%%%%%%%%%%%%%%%%%%%%%%%%%%%%%%%%%%%%%%%%%%%%%%%%%%%%%%%%%%%%%%%%%%%%%%%%%%%%%%%%
\begin{abstract}
We present decentralized collision avoidance algorithms for quadrotor swarms operating under uncertain state estimation. Our approach exploits the differential flatness property and feedforward linearization to approximate the quadrotor dynamics and performs reciprocal collision avoidance. We account for the uncertainty in position and velocity by formulating the collision constraints as chance constraints, which describe a set of velocities that avoid collisions with a specified confidence level. We present two different methods for formulating and solving the chance {\color{black}constraints}: our first method assumes a Gaussian noise distribution. Our second method is its extension to the non-Gaussian case by using a Gaussian Mixture Model (GMM). {\color{black}We reformulate the linear chance constraints into equivalent deterministic constraints, which are used with an MPC framework to compute a local collision-free trajectory for each quadrotor}. We evaluate the proposed algorithm in simulations on benchmark scenarios and highlight its benefits over prior methods. We observe that both the Gaussian and non-Gaussian methods provide improved collision avoidance performance over the deterministic method. On average, the Gaussian method requires $\sim5ms$ to compute a local collision-free trajectory, while our non-Gaussian method is computationally more expensive and requires $\sim9ms$ on average in {\color{black} scenarios with 4 agents}. 

%We present decentralized collision avoidance algorithms for quadrotor swarms operating under uncertain state estimation. Our approach exploits the differential flatness property and feedforward linearization to approximate the quadrotor dynamics and performs reciprocal collision avoidance. We account for the uncertainty in position and velocity by formulating the collision constraints as chance constraints, which describe a set of velocities that avoid collisions with a specified confidence level. We present two different methods for formulating and solving the chance constraint: our first method assumes a Gaussian noise distribution. Our second method is its extension to the non-Gaussian case by using a Gaussian Mixture Model (GMM). We reformulate the linear chance constraints into equivalent deterministic constraints on mean and covariance. Subsequently, the deterministic constraints are introduced in the MPC framework to compute a local collision-free trajectory for each quadrotor. We evaluate the proposed algorithm in simulations on benchmark scenarios and highlight its benefits over prior methods. We observe that both the Gaussian and non-Gaussian methods provide improved collision avoidance performance over the deterministic method. Further, the non-Gaussian method results in a relatively shorter path length compared to Gaussian formulations. On average, the Gaussian method requires $\sim5ms$ to compute a local collision-free trajectory, while our non-Gaussian method is computationally more expensive and requires $\sim7ms$ on average in the presence of 4 agents. 

\end{abstract}

%%%%%%%%%%%%%%%%%%%%%%%%%%%%%%%%%%%%%%%%%%%%%%%%%%%%%%%%%%%%%%%%%%%%%%%%%%%%%%%%

\section{Introduction}
Recent advances in Unmanned Aerial Vehicles (UAVs) have led to many new applications for aerial vehicles. These include search {\color{black}and} rescue, last-mile delivery, and surveillance, {\color{black}and they benefit from the small size and maneuverability of quadrotors}. Furthermore, many of these applications use a large number of quadrotors (e.g., swarms). A key issue is developing robust navigation algorithms so that each quadrotor agent avoids collisions with other dynamic and static obstacles in its environment. Moreover, in general, the quadrotors {\color{black}{need to}} operate in uncontrolled outdoor settings like urban regions, where the agents rely on onboard sensors for state estimation. In practice, onboard sensing can be noisy, which can significantly affect the collision avoidance performance.
%Recent advances in Unmanned Aerial Vehicles (UAVs) have led to many new applications for aerial vehicles. These include search \& rescue, last-mile delivery, and surveillance, {\color{black}and they benefit from the small size and maneuverability of quadrotors}. Furthermore, many of these applications use a large number of quadrotors (e.g., swarms). A key issue is developing robust navigation algorithms so that each quadrotor agent avoids collisions with other dynamic and static obstacles in its environment. Moreover, in general, the quadrotors need to operate in uncontrolled outdoor settings like urban regions, where the agents have to rely on onboard sensors for state estimation. In practice, onboard sensing can be noisy, which can significantly affect the collision avoidance performance. As a result, we need to develop appropriate collision avoidance methods. 
%making the development of probabilistic collision avoidance methods very relevant. 

Prior work on collision-free navigation is broadly classified into centralized and decentralized methods. Centralized methods~\cite{Augugliaro, Kushleyev, Preiss} plan collision-free trajectories for all agents in a swarm simultaneously, and they can also provide guarantees on smoothness, time optimality, and collision avoidance. However, due to the centralized computation, these algorithms do not scale well with the number of agents. In decentralized methods~\cite{VO, RVO, ORCA, Morgan, zhou}, each agent makes independent decisions to avoid a collision. In practice, they are scalable due to the decentralized decision making, but do not guarantee optimality or reliably handle uncertainty.
%Furthermore, each agent needs to account for sensor uncertainty for collision avoidance.

%In this paper, we present a chance constraint-based collision avoidance method for quadrotor swarm. Our method considers state and ego-motion uncertainty, solves a model predictive control MPC problem to compute a collision avoiding control input. and we show two formulations one considering Gaussian noise and the other considering random non-parametric noise. We compare both the formulations with bounding volume-based methods in the presence of \_\_\_\_\_ noise.
% \begin{figure}[t]
%     \centering
%     %\framebox{\parbox{3in}{We suggest that you use a text box to insert a graphic (which is ideally a 300 dpi TIFF or EPS file, with all fonts embedded) because, in an document, this method is somewhat more stable than directly inserting a picture.}}
%     \includegraphics[width=2.5in, height=2in]{example-image-a}
%     \caption{Example Schematic}
%     \label{fig:trajProposed}
% \end{figure}

In addition, prior work on multi-agent collision avoidance is limited to deterministic settings. These methods are mainly designed for indoor environments, where the physical evaluations are performed with a MoCap-based state estimation. On the other hand, real-world quadrotor deployment relies on onboard sensor data, which can be noisier. For example, depth cameras are widely used in robotics applications, {\color{black}{but the estimated depth values may have errors}} due to lighting, calibration, or object surfaces~\cite{khoshelham2012accuracy,park2020efficient}. Some of the simplest techniques consider zero-mean Gaussian uncertainty by enlarging the agent's bounding geometry in relation to the variance of uncertainty~\cite{Snape, Kamel}.
%Bounding volume expansion~\cite{Snape,Kamel} presents a robust method to consider Gaussian uncertainty by enlarging the agent's bounding geometry in relation to the variance of uncertainty. But,
However, these methods tend to over-approximate the collision probability, resulting in conservative navigation schemes~\cite{Zhu}. Other uncertainty algorithms are based on chance-constraint methods~\cite{Zhu, PRVO}. {\color{black}These algorithms} are less conservative in practice, but assume simple agent dynamics or {\color{black}are limited to simple scenarios}. %They are less conservative in practice, but assume simple agent dynamics or are limited to very simple scenarios.

{\color{black}{\subsection{Main Results:}}}
We present a decentralized, probabilistic collision avoidance method (SwarmCCO) for quadrotor swarms operating in dynamic environments. Our approach builds on prior techniques for multi-agent navigation based on reciprocal collision avoidance~\cite{ORCA, DCAD}, and we present efficient techniques to perform probabilistic collision avoidance by chance-constrained optimization (CCO). We handle the non-linear quadrotor dynamics using flatness-based feedforward linearization. The reciprocal collision avoidance constraints are formulated as chance constraints and {\color{black}{combined with the MPC (Model Predictive Control) framework}}. %{\color{red}{CAN YOU MOTIVATE WHY YOU WANT TO USE CHANE CONSTRAINTS TO MODEL UNCERTAINTY? WHAT IS SO INTRINSIC ABOUT THESE CHANCE CONSTRAINTS THAT MAKES THEM A GOOD CANDIDATE FOR THIS APPLICATION?}} We describe two algorithms for formulating and solving the CCO: 
\begin{itemize}
    \item {\color{black}Our} first algorithm assumes a Gaussian noise distribution for the state uncertainties and reformulates the {\color{black}collision avoidance} chance constraints as a set of deterministic second-order cone constraints.
                  
    \item {\color{black}Our} second algorithm is designed for non-Gaussian noises. {\color{black}We use a Gaussian Mixture Model (GMM) to approximate the noise distribution and replace each collision avoidance constraint using a second-order cone for each Gaussian components.} The cone constraints for the individual Gaussian components are related to the GMM's probability distribution {\color{black}by using an additional constraint based on GMM's mixing coefficient.}%through an additional constraint using the GMM's mixing coefficient.%. We use an efficient branch-and-bound method  {~\cite{ref}} to solve.}
\end{itemize}
%For simplicity, we assume a Gaussian uncertainty distribution and we reformulate the chance constraint as a deterministic constraint on mean and covariance. 
%WHAT ARE THE OVERALL BENEFITS OF YOUR APPROACH IN TERMS OF HANDLING UNCERTAINTY? TALK ABOUT SOME RELATIVE BENEFITS OF BOTH GAUSSIAN AND NON-GAUSSIAN METHODS? IN WHAT SITUATION IS THE GAUSSIAN FORMULATION USEFUL, AND WHAT APPROACHES ARE THE NON-GAUSSIAN USEFUL?
We evaluate our probabilistic methods (SwarmCCO) in simulated environments with a large number of quadrotor agents. We compare our probabilistic method's performance with the deterministic algorithm~\cite{DCAD} in terms of the path length, time to goal, and {\color{black}the number of collisions}. We observe {\color{black}that} both our Gaussian and non-Gaussian methods result in {\color{black}fewer} collisions in the presence of noise. {\color{black}Our average computation time is} $\sim5ms$ per agent for our Gaussian method and $\sim9ms$ per agent for the non-Gaussian method in {\color{black}scenarios with $4$ agents}. {\color{black}The non-Gaussian method is computationally expensive compared to the Gaussian method, but the non-Gaussian method provides improved performance in terms of shorter path lengths, and satisfying collision avoidance constraints (Section V-D)}. Hence, the non-Gaussian method {\color{black}tends to} offer better performance in constricted regions due to better approximation of noise, {\color{black}where the Gaussian method may result in an infeasible solution}.

{\color{black}The paper is organized as follows.} Section II summarizes the recent relevant works in probabilistic collision avoidance. Section III provides a brief introduction to DCAD~\cite{DCAD} and ORCA~\cite{ORCA} chance-constraints. {\color{black}In Section IV, we present our algorithms and describe the chance constraint formulation}. In Section V, we present our results and compare the performance with other methods. In Section V, we present our results and compare the performance with other methods. In Section VI, we summarize our major contributions, results, and present the limitations and future work.
%evaluations and results of our two algorithms. The first algorithm assumes a Gaussian noise distribution for the state uncertainties and reformulates the ORCA chance constraints as a set of deterministic second-order cone constraints.

\section{Previous Work}
In this section, we provide a summary of the recent work on collision avoidance and trajectory planning under uncertainty.

\subsection{Decentralized Collision Avoidance with Dynamics}
Decentralized collision avoidance methods~\cite{VO,RVO,ORCA,Morgan,AVO} compute the paths by locally altering the agent's path based on the local sensing information and state estimation. Velocity Obstacle (VO)~\cite{VO} methods such as RVO~\cite{RVO} and ORCA~\cite{ORCA} provide decentralized collision avoidance for agents with single-integrator dynamics. This concept was extended to double integrator dynamics in the AVO algorithm~\cite{AVO} and used to generate $n^{th}$ order continuous trajectories in \cite{cnco}. Berg et al.~\cite{LQG} and Bareiss et al.~\cite{LQR} proposed control obstacles for agents with linear dynamics. Moreover, the authors demonstrated the algorithm on quadrotors by linearizing about the hover point. %\sout{However, hover point linearization is valid only for small attitude deviations about the hover point~\cite{hoverlimitation}}.
Cheng et al. \cite{MPCORCA} presented a variation by using ORCA constraints on velocity and a linear MPC to account for dynamics. Morgan et al.~\cite{Morgan} described a sequential convex programming (SCP) method for trajectory generation. However, SCP methods can be computationally expensive for rapid online replanning. Most of these methods have been designed for deterministic settings. Under imperfect state estimation and noisy actuation, the performance of deterministic algorithms may not be reliable and can lead to collisions {{\cite{PRVO}}}. Hence, we need probabilistic collision avoidance methods for handling uncertainty.

\subsection{Uncertainty Modeling}
Snape et al.~\cite{Snape} extended the concept of VO to address state estimation uncertainties using Kalman filtering and bounding volume expansion for single-integrator systems. That is, the agent's bounding polygon is enlarged based on the co-variance of uncertainty. Kamel et al.~\cite{Kamel} proposed an N-MPC formulation for quadrotor collision avoidance and used the bounding volume expansion to address sensor uncertainties. DCAD~\cite{DCAD} presented a collision avoidance method for quadrotors using ORCA and bounding volume expansion. Bounding volume expansion methods retain the linearity of ORCA constraints; hence, they are fast but tend to be conservative. They do not differentiate samples close to the mean from those farther away from the mean~\cite{Zhu, arxivChance}. Hence, they can lead to infeasible solutions in dense scenarios~\cite{Zhu}. Angeris et al.~\cite{angeris2019fast} accounted for uncertainty in estimating a neighbor's position using a confidence ellipsoid before computing a safe reachable set for the agent.

In contrast to bounding volume methods, \cite{Zhu,B-UAVC} modeled the stochastic collision avoidance as a chance-constrained optimization. These techniques assumed a Gaussian noise distribution for the position and transformed the chance constraints to deterministic constraints on mean and co-variance of uncertainty. Gopalakrishnan et al.~\cite{PRVO} presented PRVO, a probabilistic version of RVO. PRVO assumed a Gaussian noise distribution and used Cantelli's bound to approximate the chance constraint. However, PRVO considers simple single-integrator dynamics. Jyotish et al.~\cite{NagaJyotish} extended PRVO to non-parametric noise and formulated the CCO problem as matching the distribution of PVO with a certain desired distribution
using RKHS embedding for a simple linear dynamical system. However, this method is computationally expensive and requires about $0.2s$ to compute a suitable velocity in the presence of 2 neighbors. 

There is also considerable literature on probabilistic collision detection to check for collisions between noisy geometric datasets~\cite{rusu2009real,lee2013sigma,du2011probabilistic,park2016fast,park2020efficient}. They are applied on point cloud datasets and used for trajectory planning in a single high-DOF robot, but not for multi-agent navigation scenarios.
%\vspace*{0.05in}\\

%In this work, we extend the DCAD algorithm \cite{DCAD} to chance constraint optimization. Our method draws from \cite{PRVO,Zhu,Blackmore} to formulate the ORCA collision avoidance constraint as a chance constraint, which under the assumption of a Gaussian uncertainty distribution is reformulated as a deterministic equation on the mean and variance. Further, we also present a method for non-gaussian uncertainty distribution by using a GMM. {\color{red}Chance constraints are in general intractable to solve~\cite{PRVO}, the Gaussian/GMM assumption simplifies the problem to obtain real-time solution of the optimization problem (Section IV).}

\section{Background and Problem Formulation}
\begin{table}[t]
 \caption{\label{tab:notation} Notation and symbols.}
\begin{center}
\renewcommand{\arraystretch}{1.1}
\resizebox{\columnwidth}{!}{%
 \begin{tabular}{|p{1.0cm}|p{6.5cm}|} 
 \hline
 \textbf{Notation} & \textbf{Definition} \\ 
 \hline\hline
  $\mathcal{W}$ & World Frame defined by unit vectors $\mathbf{x_W}$, $\mathbf{y_W}$, and $\mathbf{z_W}$ along the standard X, Y and Z axes \\ 
 \hline
 $\mathcal{B}$ & Body Frame attached to the center of mass, defined by the axes $\mathbf{x_B}$, $\mathbf{y_B}$, and $\mathbf{z_B}$ \\
 \hline
 $\mathbf{r}_{i}$ & 3-D position of $i^{th}$ quadrotor given by $[r_{i,x}, r_{i,y}, r_{i,z}]$ \\
% \hline
% $\mathbf{v}$ & Velocity of the quadrotor given by $[\dot{x}, \dot{y}, \dot{z}]$\\
 \hline
 $\mathbf{v}_i, \mathbf{a}_i$ & 3-D Velocity and Acceleration of $i^{th}$ quadrotor given by $[v_{i,x}, v_{i,y}, v_{i,z}]$ and $[a_{i,x}, a_{i,y}, a_{i,z}]$, respectively\\
 \hline
 ${R}_i$ & Radius of agent $i$'s enclosing sphere\\
 \hline
 $\phi, \theta, \psi$ & Roll, pitch and yaw of the quadrotor\\
 \hline
 $\mathbf{R}$ & Rotation matrix of quadrotor body frame ($\mathcal{B}$) w.r.t world frame ($\mathcal{W}$)\\% i.e., {\color{red} $\mathcal{B}$ = $\mathbf{R}\mathcal{W}$}\\
 \hline
 %$\mathbf{J}$ & Quadrotor inertia matrix given by $diag(J_{xx}, J_{yy}, J_{zz})$\\
 %\hline
 $\mathbf{T}$ & Net thrust in body fixed coordinate frame\\
 \hline
 $\mathbf{m}_q$ & Mass of quadrotor\\
 \hline
 ${\boldsymbol{\omega}}$ & Angular velocity in body fixed coordinate frame given by $[p,q,r]$\\
 \hline
 %$f^{M_{j}^{i}} (.)$ & Collision avoidance velocity constraint given by RVO (when M is RVO) and ORCA (when M is ORCA)\\
 %\hline
 $\mathbf{v}_{i}^{orca}$ & Collision avoiding velocity for agent $i$\\
 \hline
  $\color{black}{f^{orca_i^j}}$ & ORCA \color{black}{plane constraint given by  $\mathbf{m}^T\mathbf{v}^{orca}_{i}-b \geq 0$. $\mathbf{m}$ and b are functions of the agents trajectory.}\\ 
 \hline
 ${\color{black}{\mu_m, \sigma_m}}$ & Mean and standard deviation of a variable `{\color{black}{m}}'\\
 \hline
  $\color{black}{P(x)}$ & { $\text{\color{black}Probability of x}$}\\ 
 \hline 
 ${\color{black}{\boldsymbol{\mu}_m, \Sigma_m}}$ & Mean and covariance of a vector
 `{\color{black}{m}}'\\
 %$\mathbf{R}$ & Rotation matrix of ($\mathcal{B}$) w.r.t ($\mathcal{W}$). $\mathbf{R}_{ij}$ represents element in the $i^{th}$ row and $j^{th}$ column.\\% i.e., {\color{red} $\mathcal{B}$ = $\mathbf{R}\mathcal{W}$}\\
 %\hline
 %$\mathbf{J}$ & Quadrotor inertia matrix given by $diag(J_{xx}, J_{yy}, J_{zz})$\\
 %\hline
 %$T$ & Net thrust in body fixed coordinate frame\\
 %\hline
 %$m$ & Mass of quadrotor\\
 %\hline
 %${\boldsymbol{\omega}}$ & Angular velocity in body fixed frame given by $[p,q,r]$\\
 \hline
% $\boldsymbol{\tau}$ & {Torque in body-fixed coordinate frame given by $[\tau_1, \tau_2, \tau_3]$}\\
% \hline 
 $\mathbf{x}, \mathbf{u}$ &{Quadrotor \color{black}{flat state and flat control} input}\\
 \hline
 %$\mathbf{u}$ & Control input to the quadrotor\\
 %\hline
 %$\mathbf{u_c}$ & Input to the inner loop controller\\
 %\hline
 %$\mathbf{ORCA^\tau_{A|B}}$ & Collision avoiding velocity set for agent A induced by agent B over a time horizon $\tau$\\
%\hline
\end{tabular}
}
\end{center}
\vspace{-10pt}
\end{table}
This section discusses the problem statement and gives an overview of various concepts used in our approach. Table \ref{tab:notation} summarizes the symbols and notations used in our paper.
\subsection{Problem Statement}
We consider $\mathrm{N}$ agents occupying a workspace $\mathcal{W}\subseteq\mathbb{R}^3$. Each agent $i \in \left\{1,2,...,\mathrm{N}\right\}$ is modeled with non-linear quadrotor dynamics, as described in {\cite{DCAD}}. For simplicity, {\color{black}each} agent's {\color{black}geometric representation} is approximated {\color{blue}as} a sphere of radius $R$. We assume that each agent knows its neighbor's position and velocity either through {\color{black}inter-agent communication or visual sensors, and this information is not precise}. No assumption is made on the nature of the uncertainty distribution {\color{black}in this information}; hence, the random variables are assumed to be non-parametric (i.e. they are assumed to follow no particular family of probability distribution). %A sample noise distribution is shown in Fig.~\ref{fig:errorDist}. []
Our algorithm approximates the distribution using a {\color{black}Gaussian Mixture Model} GMM. %The uncertainty in the position and velocity data is assumed to follow a Gaussian distribution, and its mean and covariance are estimated using a Kalman filter (Section \ref{subsec:uncertainty}).

At any time instant, two agents $i$ and $j$, where $i, j \in \left\{1,2,...,\mathrm{N} \right\}$, $i\neq j$, are said to be collision-free if their separation is greater than sum of their bounding sphere radii. That is, {\color{black}$\left\Vert \mathbf{r}_{i} - \mathbf{r}_{j} \right\Vert_2 \ge R_i + R_j$.} Since the position {\color{black}and velocity are random variables}, collision avoidance is handled {\color{black}using} a stochastic method based on chance constraints.
%We consider $\mathrm{N}$ agents occupying a workspace $\mathcal{W}\subseteq\mathbb{R}^3$. Each agent $i \in \left\{1,2,...,\mathrm{N}\right\}$ is modeled with non-linear quadrotor dynamics, as described in {\cite{DCAD}}. For simplicity, the agent's geometry is approximated by a sphere of radius $R$. We assume that each agent knows its neighbor's position and velocity either through perception or communication. No assumption is made on the nature of the uncertainty distribution; hence, the random variables are assumed to be non-parametric (i.e. they are assumed to follow no particular family of probability distribution). A sample noise distribution is shown in Fig.~\ref{fig:errorDist}. Our algorithm approximates the distribution using a GMM. %The uncertainty in the position and velocity data is assumed to follow a Gaussian distribution, and its mean and covariance are estimated using a Kalman filter (Section \ref{subsec:uncertainty}).

At any time instant, two agents $i$ and $j$, where $i, j \in \left\{1,2,...,\mathrm{N} \right\}$, $i\neq j$, are said to be collision-free if their separation is greater than sum of their bounding sphere radii. That is,
\begin{equation*}
        \left\Vert \mathbf{r}_{i} - \mathbf{r}_{j} \right\Vert_2 \ge R_i + R_j.
\end{equation*}
Since the position is a random variable, collision avoidance is handled through a stochastic method based on chance constraints.
%The proposed method is based on ORCA~\cite{ORCA} and DCAD~\cite{DCAD}, which are discussed in the following subsections. 

\subsection{ORCA}
ORCA~\cite{ORCA} is a velocity obstacle-based method that computes a set of velocities that are collision-free. Let us consider the RVO equation as given in~\cite{PRVO}.

\begin{equation}
    f^{RVO_j^i}(\mathbf{r}_{i},\mathbf{r}_{j},\mathbf{v}_{i},\mathbf{v}_{j}, \mathbf{v}^{rvo}_{i})\geq 0,\\ 
\end{equation}
\begin{equation}
    f^{RVO_{j}^{i}} (.) = \Vert \mathbf{r}_{ij}\Vert^{2}-\frac{((\mathbf{r}_{ij})^{T}(2\mathbf{v}_{i}^{rvo}-\mathbf{v}_{i}-\mathbf{v}_{j}))^{2}}{\Vert 2\mathbf{v}_{i}^{rvo}-\mathbf{v}_{i}-\mathbf{v}_{j}\Vert^{2}}-(R_{ij})^{2},\\
    \label{frvo}
\end{equation}

\begin{equation}
    \mathbf{r}_{ij}=\mathbf{r}_{i}-\mathbf{r}_{j}, \\ R_{ij} = R_{i} + R_{j}.
\end{equation}
Since DCAD considers linear constraints given by ORCA, we construct the ORCA constraints from Eqn.~(\ref{frvo}) by linearizing the function about an operating point. In our case, the operating point is chosen as a velocity on the surface of truncated VO cone closest to the relative velocity between the two agents. The ORCA constraint (linearized equation) has the following form:
\begin{equation}\label{forcaE}
    f^{ORCA_{j}^{i}} (.) =
        {\mathbf{m}^T\mathbf{v}^{orca}_{i}-b \geq 0.}
        %\mathbf{a}^T\mathbf{v}^{rvo}_{i}-b \leq 0.
\end{equation}
Here, $\mathbf{v}^{orca}_{i}$ is any velocity in the half-space of collision-free velocities. Eqn.~\ref{forcaE} is used to construct the chance constraint, which is detailed in Section IV.

\begin{figure}[t]
    \centering
    %\framebox{\parbox{3in}{We suggest that you use a text box to insert a graphic (which is ideally a 300 dpi TIFF or EPS file, with all fonts embedded) because, in an document, this method is somewhat more stable than directly inserting a picture.}}
    \includegraphics[width=0.99\linewidth]{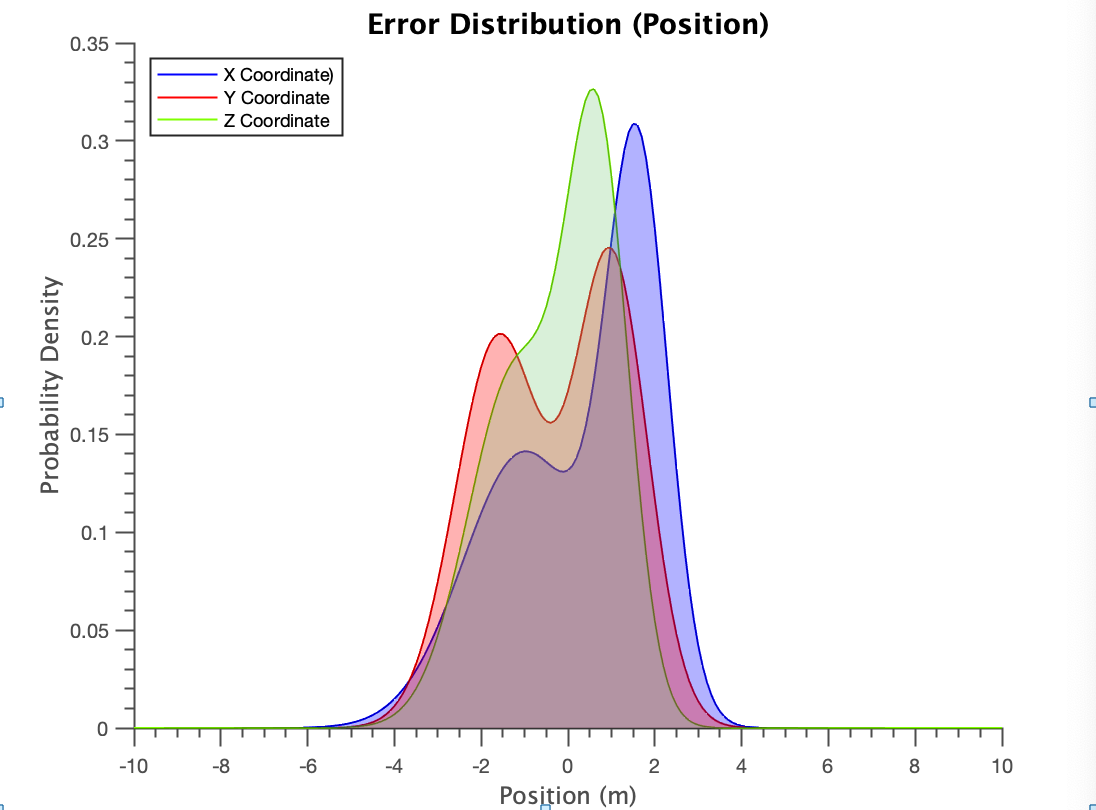}
    \caption{A sample additive non-Gaussian noise distribution added to the 3D position of the agent. The distribution is generated using a 5 component GMM.}
    \label{fig:errorDist}
\end{figure}

{\color{black}{\subsection{Differential Flatness}
A non-linear system $\dot{\mathbf{x}}_o = f(\mathbf{x}_o,\mathbf{u}_o)$ is differentially flat if a set $\zeta = [\zeta_1, \zeta_2, ..., \zeta_m] $ of differentially independent components and their derivatives can be used to construct the system state space and control inputs~\cite{DCAD}. Here, $\mathbf{x}_o$ and $\mathbf{u}_o$ represent the state and control input of the non-linear system.\\

Given a non-linear, differentially flat system and a smooth reference trajectory in $\zeta$ (denoted $\zeta_{ref}$), the nonlinear dynamics can be represented as a linear flat model ($\dot{\mathbf{x}} = A\mathbf{x} + B\mathbf{u}$) using feedforward linearization. The vectors $\mathbf{x}$ and $\mathbf{u}$ represent the flat state and flat control input respectively. From the definition of differential flatness, a reference state ($\mathbf{x}_{ref}$) and control input ($\mathbf{u}_{ref}$) can be constructed using $\zeta_{ref}$.

Formally, a differentially flat, non-linear system $\dot{\mathbf{x}}_o = f(\mathbf{x}_o,\mathbf{u}_o)$ can be feedforward linearized into the following linear system
$$
\dot{\mathbf{x}} = A\mathbf{x} + B\mathbf{u}
$$
$$
\mathbf{u} = g(\mathbf{x}, \mathbf{u}_o, \dot{\mathbf{u}}_o, ... , \mathbf{u}_o^\sigma)
$$
provided a nominal control input $\mathbf{u}_o^*$ computed from ($\mathbf{x}_{ref}$) and ($\mathbf{u}_{ref}$) is applied to the non-linear system, and the initial condition of the reference trajectory is consistent with the current state of the non-linear system. That is, when the following two conditions are satisfied.
$$
\mathbf{u}_o = g^{-1}(\mathbf{x}_{ref}, \mathbf{u}_{ref})
$$
$$
\mathbf{x}(0) = \mathbf{x}_{ref}(0) 
$$
Here, $g$ represents some non-linear mapping and $\sigma$ represents the highest power of $\mathbf{u}_o$ required to construct the flat input $\mathbf{u}$.
A quadrotor is a differential flat system with $\zeta = [r_x, r_y, r_z, \psi]$~\cite{mellinger}. We use the linear flat model in the MPC problem (as shown in Equation ~\ref{eqn:optimization}), and compute $\mathbf{x}_{ref}$ and $\mathbf{u}_{ref}$, which are used to compute a nominal control input $\mathbf{u}_o^*$ for the quadrotor using an inverse mapping (section III-D).
}}

\subsection{DCAD}\label{sec:DCAD}
Decentralized Collision Avoidance with Dynamics (DCAD)~\cite{DCAD} is a receding horizon planner for generating local, collision-free trajectories for quadrotors. DCAD exploits the differential-flatness property of a quadrotor to feedforward linearize the quadrotor dynamics and uses linear MPC and ORCA constraints to plan a collision-free trajectory in terms of differentially-flat states. Further, DCAD uses an inverse mapping to account for the non-linear quadrotor dynamics by transforming the flat control inputs into inner loop controls. DCAD accounts for uncertainty in state estimation by assuming Gaussian noise and uses bounding volume expansion to account for the uncertainties. Our method differs from DCAD in posing the ORCA linear constraints as chance constraints and performing a chance-constrained optimization to compute a collision-avoiding input for the quadrotor. Further, in this work we use a flat state space of 7 states given by, $\mathbf{x} = [\mathbf{r}_i, \mathbf{v}_i, \psi].$ The flat control input is given by, $\mathbf{u} = [\mathbf{a}_i, \dot{\psi}]$. The feedforward linearized dynamics model is used in our optimization problem~\ref{eqn:optimization}.%We highlight the benefits in terms of highlighting uncertainty in Section V.

\noindent{\bf{Quadrotor Model:}}
The quadrotor state space and the control input are given by
\begin{eqnarray}
{\bf{x}_o} = [x, y, z, \dot{x}, \dot{y}, \dot{z}, \phi, \theta, \psi, p, q, r],\\
{\bf{u}_o} = [T, {\phi}, {\theta}, \dot{\psi}].
\end{eqnarray}
The quadrotor dynamics we consider is same as in prior literature~\cite{mellinger, ferrin}. The states and control input are similar to~\cite{ferrin}.
    The quadrotor dynamics can be represented by the following set of equations:
    \begin{eqnarray}
    \label{eqn:rfot=v}
    \mathbf{\dot{r}} = \mathbf{v},\\
    \label{eqn:accel}
    \mathbf{m\bar{a} = -mgz_{W} + Tz_{B}},\\
    \label{eqn:rdot}
    \mathbf{\dot{R} = R \times \boldsymbol{\omega}^T},\\
    \label{eqn:tau}
    \boldsymbol{\dot{\omega}} = \mathbf{j^{-1}}[-\boldsymbol{\omega} \times \mathbf{j}\boldsymbol{\omega} + \boldsymbol{\tau}].
    \end{eqnarray}

We consider the flat output set given by $\boldsymbol{\zeta} = [x,y,z,\psi]$ similar to~\cite{mellinger}.
In SwarmCCO, the flat state ($\mathbf{x}$) and flat input ($\mathbf{u}$) are given by,
$$\mathbf{x} = [r_x, r_y, r_z, v_x, v_y, v_z, \psi],$$
$$\mathbf{u} = [a_x, a_y, a_z, \dot{\psi}].$$
The system dynamics in flat states is linear and hence is used as the agent dynamics in the optimization (MPC) problem (2). This results in faster computation than when compared to using the non-linear quadrotor model. Since $\mathbf{x}$ and $\mathbf{u}$ represent the flat state and input in the optimization problem (2) we need to transform the output from (2) to the original quadrotor input $\mathbf{u_o}$. The optimization problem (2) corresponds to a model predictive control (MPC) problem has an output given by optimized flat control inputs for $N-1$ time steps. Here, N is the time horizon of the MPC problem. Considering the mass of the quadrotor as $m_q$, we can compute the quadrotor control input ($\mathbf{u_o}$) from the flat inputs ($\mathbf{u}$) by,
    $$T = m_q\sqrt{a_x^2 + a_y^2 + a_z^2}$$
$$\begin{bmatrix} z_1\\ z_2\\ z_3 \end{bmatrix} = Rot(\psi)\begin{bmatrix}
    a_x\\ a_y\\ a_z\\
    \end{bmatrix}
     \frac{m_q}{-T}$$
    $$\phi = \sin{(-z_2)}$$
    $$\theta = \tan{(z_1/z_3)}$$
    $$\psi = \dot{\psi}\cos{(\theta)}\cos{(\psi)} - \dot{\theta}\sin{(\phi)}$$
    
Here, $Rot$ denotes the rotation matrix. Thus, we can transform flat inputs $\mathbf{u} = [a_x, a_y, a_z, \dot{\psi}]$ to quadrotor control input $\mathbf{u_o}=[T, \phi, \theta, \psi]$. This set of equations transforming $\mathbf{u}$ to $\mathbf{u_o}$ is the inverse mapping.
% \begin{figure}[t]
%     \centering
%     \includegraphics[width=0.8\linewidth]{example-image-a}
%     \caption{DCAD}
%     \label{fig:DCAD}
% \end{figure}

\subsection{Chance Constraints}
Chance-constrained optimization is a technique for solving optimization problems {\color{black}with} uncertain variables~\cite{cooperCCP,Prekopa}. %It formulates the optimization problem such that the probability of satisfying a given constraint is above a certain confidence level (or probability). 
A general formulation for chance-constrained optimization is given {as}
\begin{equation*}
\begin{aligned}
& \underset{}{\text{minimize}} \quad
& & \gamma \\
& \textrm{subject to} \quad
& & {\textnormal{P}(f(x)\le \epsilon)\ge \delta.}\\
\end{aligned}
%\label{eqn:optimization}
\end{equation*}
Here, $\gamma$ is the objective function for the optimization. $f(x)$ is the constraint {on} the random variable $x$. $f(x)$ is said to be satisfied when $f(x)\le \epsilon$. Since x is a random variable, the constraint $f(x)\le \epsilon$ is formulated as the probability of satisfying of the constraint. That is, {\color{black}$\textnormal{P}(f(x)\le \epsilon)\ge \delta$} is the chance constraint and is said to be satisfied when the probability of satisfying the constraint  $f(x)\le \epsilon$ is over a specified confidence level, $\delta$.

%BRIEFLY MENTION WHY AND HOW DO YOU MODEL UNCERTAINTY AS CHANCE CONSTRAINTS? WHY IS CHANCE CONSTRAINT A GOOD CHOICE TO MODEL UNCERTAINTY, INCLUDING GAUSSIAN OR GMM FORMULATIONS?

% \subsection{Uncertainty Propagation}\label{subsec:uncertainty}
% {\color{black}{To account for a quadrotor’s imperfect sensing, we use a Kalman filter to estimate the mean and covariance of the position and velocity of the agent. The agent’s process and observation model is assumed to be as follows,
% \begin{align*}
%     x_{k+1} = Ax_k +Bu_k +w_k\\
%     y_k = Cx_k +v_k
% \end{align*}
% Here $x_{k}$ is the state and $u_{k}$ is the control input at time step k. $w_k$ and $v_k$ are the process and measurement noise respectively and are assumed to be Gaussian, zero-mean white-noise. The state transition matrix A, matrix B and C are same as in the linear flat model we compute for the
% quadrotor. The mean and the co-variance for the states are computed through Kalman predict and update cycle.}}
\section{SwarmCOO: Probabilistic Multi-agent Collision Avoidance}
\begin{figure*}[t]
   \centering
   \begin{subfigure}[b]{0.32\textwidth}
   \includegraphics[height=1.6in,width=0.90\linewidth]{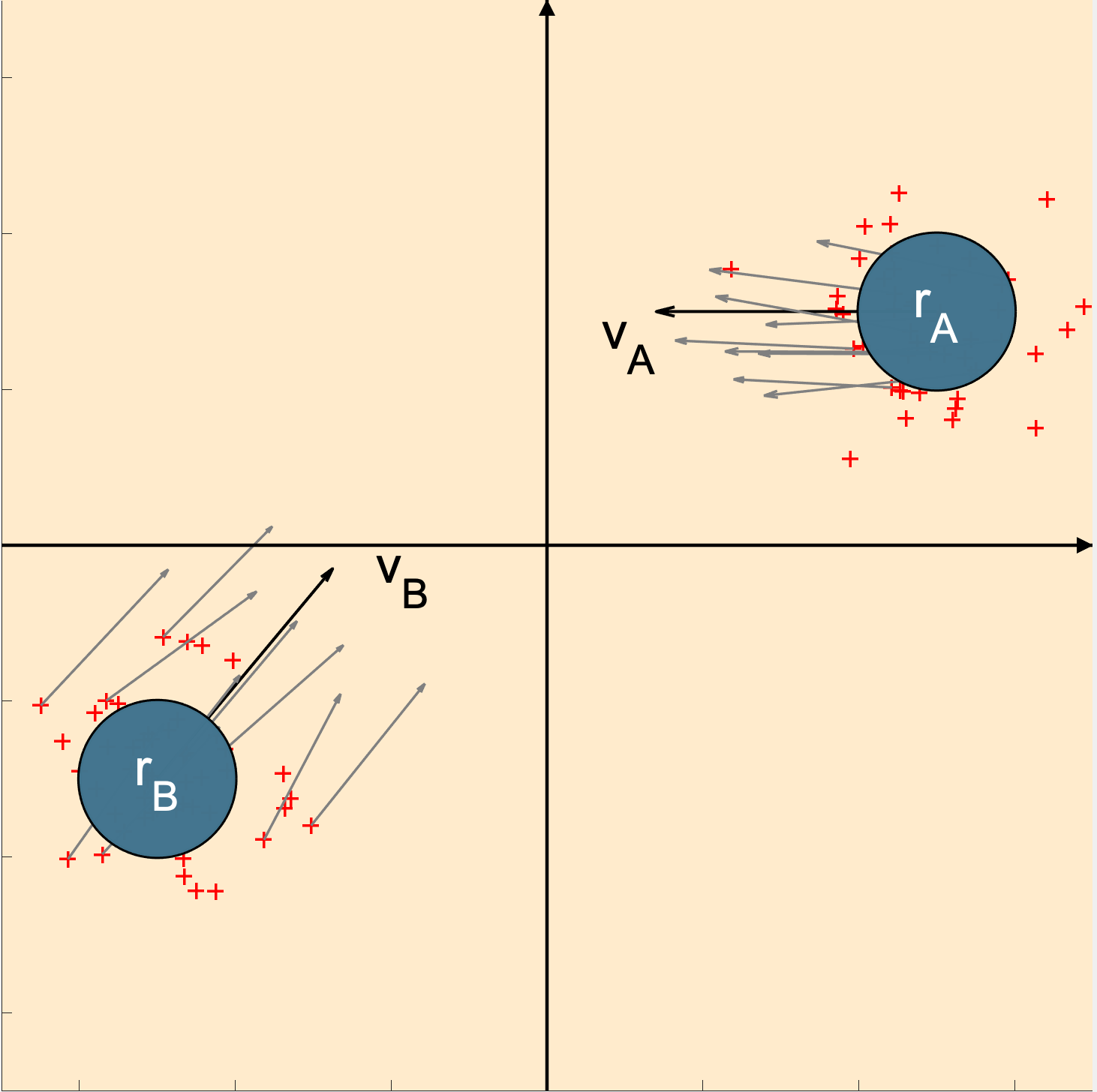}
   %{New_be.jpg}
   \caption{Two-robot scenario}
   \label{fig:New_traj}
   \end{subfigure}
   %\hfill
   \begin{subfigure}[b]{0.32\textwidth}
   %Screen Shot 2020-02-27 at 12.44.34 PM
   \includegraphics[height=1.6in,width=0.90\linewidth]{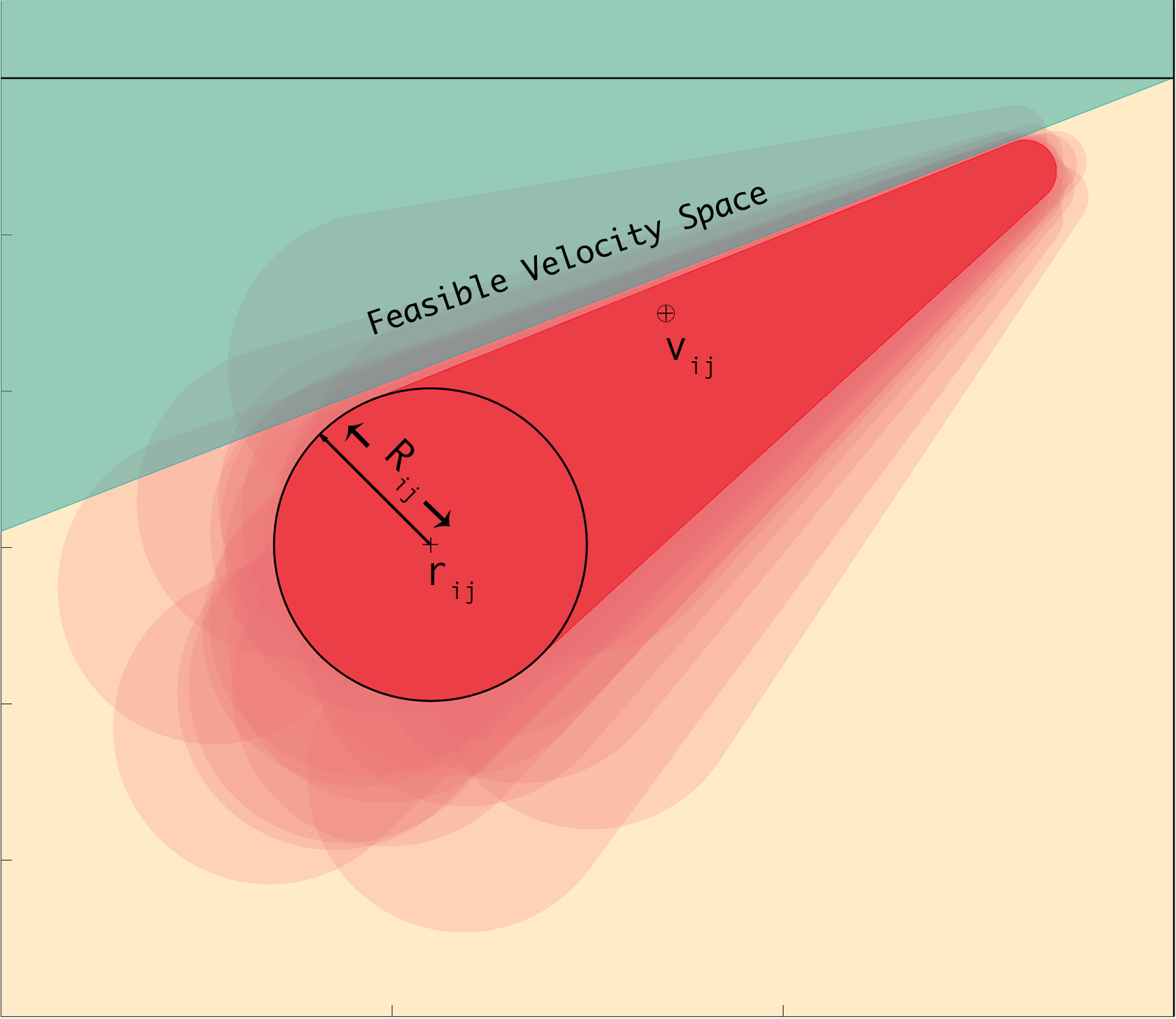}
   %{orca_be.jpg}
   \caption{Deterministic ORCA}
   \label{fig:Orca_traj}
   \end{subfigure}
   %\hfill
   \begin{subfigure}[b]{0.32\textwidth}
   %screen Shot 2020-02-27 at 1.39.42 PM
   \includegraphics[height=1.6in,width=0.90\linewidth]{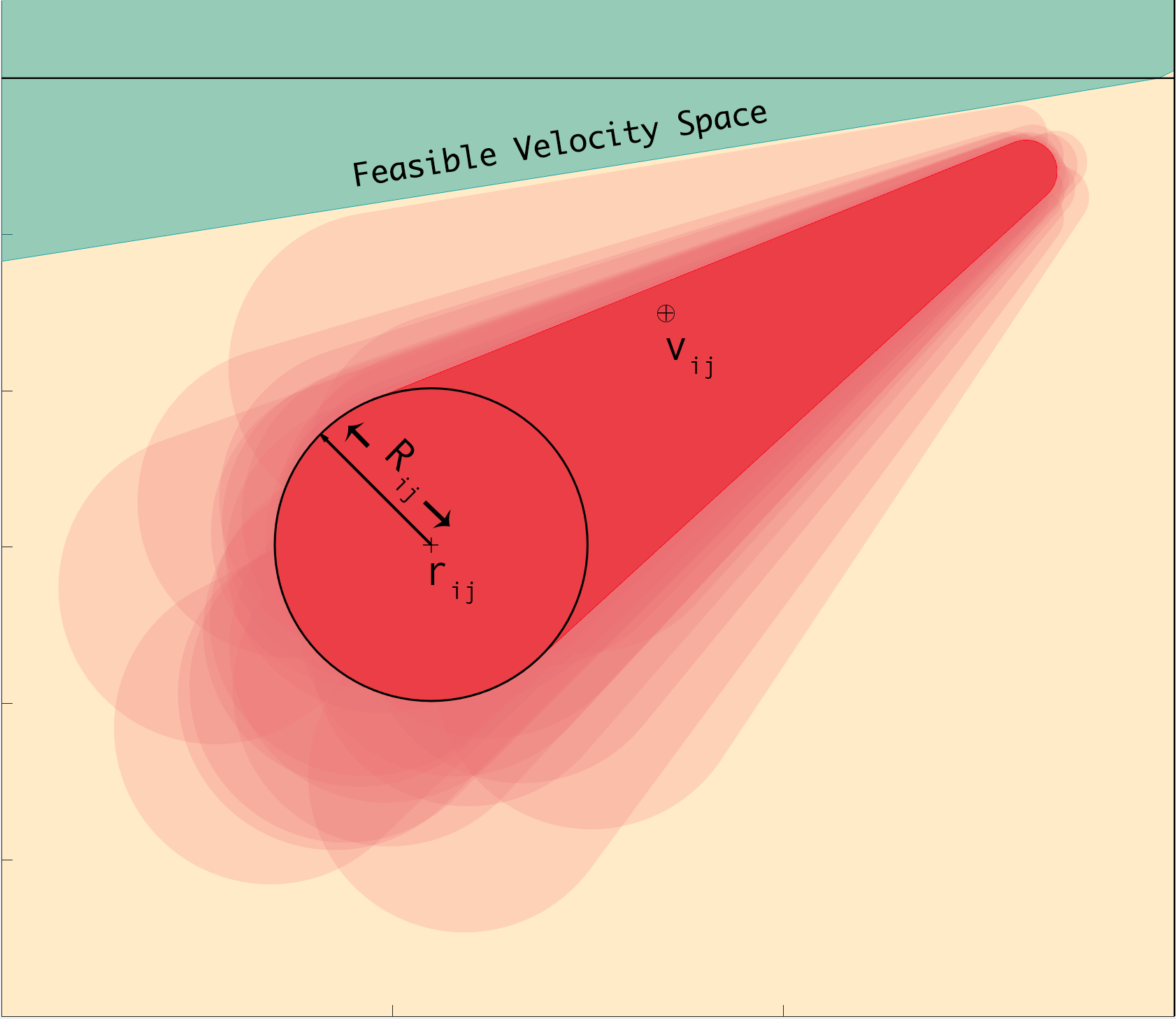}%{Avo_be.jpg}
   \caption{Chance Constrained ORCA}
   \label{fig:Avo_traj}
   \end{subfigure}
   \caption{Deterministic and Probabilistic collision avoidance between two agents. (a) Two circular agents with mean positions $r_i$ and $r_j$ and their respective mean velocities $v_i$ and $v_j$ (indicated by the black arrow) are shown. The red `{\color{black}{+}}' markers indicate position samples from the position's uncertainty distribution{\color{black},} and the {\color{black}{grey arrows}} indicate velocity samples from the velocity's uncertainty distribution. (b) The {\color{black}{\textit{darker}}} red cone represents the Velocity Obstacle (VO) constructed using the mean position and velocity, while the {\color{black}{\textit{lighter}}} or {\color{black}{\textit{translucent}}} red cones represent the VOs constructed from the position and velocity samples from the uncertainty distribution. The {\color{black}{green}} region is the feasible region for the relative velocity computed using ORCA. As can be seen, the {\color{black}green} regions overlap with parts of VOs constructed using the position and velocity samples, which can lead to {\color{black}collisions}. (c) {\color{black}Feasible relative velocity set computed using the chance constraint ORCA method avoids a major portion of the VO samples.}}
   \label{fig:sampleScenario}
   \vspace{-8pt}
\end{figure*}
In this section, we describe our MPC optimization problem and summarize our Gaussian and non-Gaussian chance constraint formulations for collision avoidance. Fig.~\ref{fig:sampleScenario} {\color{black}highlights a 2D scenario for collision avoidance} between two agents with state uncertainty. The figure shows a distribution of Velocity Obstacle (VO) cones constructed for the given position and velocity distribution. We observe that the collision-free relative velocity set computed using deterministic ORCA overlaps with a portion of this VO distribution. Hence, a velocity chosen from this set can result in a collision. In contrast, {\color{black}our formulation based on} chance constraints results in a feasible relative velocity set that has a higher probability of being collision-free.
\subsection{\color{black}MPC Optimization Setup}
We use a receding horizon planner to generate the collision-free trajectories for each quadrotor agent. {\color{black}The underlying optimization framework} is common to both our Gaussian and non-Gaussian SwarmCCO formulations. {\color{black}Our} Gaussian and non-Gaussian methods differ in the formulation for collision avoidance chance constraints, i.e. the constraint $\text{P}\left({\color{black}{\mathbf{m}^T {\mathbf{v}^{orca}_{i}-b\le0}}}\right) \le \delta$ in the optimization is formulated differently~(\ref{subsec:ccf}). {\color{black}Each pair of agents has a constraint $\text{P}\left({\color{black}{\mathbf{m}^T {\mathbf{v}^{orca}_{i}-b\le0}}}\right) \le \delta$. That is, if an agent has 5 neighbors, there would be 5 chance constraints in Eqn. (\ref{eqn:optimization}). The variables $\mathbf{m}$ and $b$ for each constraint would depend on the relative positions and velocities between that pair of agents. For clarity, we have considers a single neighbor case in this section.} {\color{black}Each quadrotor agent} computes the collision avoidance constraints at each time step and plans a trajectory for the next $N$ time steps. The trajectory is re-planned continuously to account for the changes in the environment.

{\color{black}In our optimization formulation (given below)}, ``$N$" represents the prediction horizon. The weight matrices{\color{black},} Q and R{\color{black},} prioritize between trajectory tracking error and the control input, respectively. $\mathbf{x}_k$ represents the state of the agent at time step k, and the matrix A and B are the system matrices for the linearized quadrotor model. Velocity and acceleration are constrained to maximum values{\color{black}, $v_{max}$ and $a_{max}$, respectively,} and are realized using box constraints.
\begin{equation}
\begin{aligned}
& \underset{}{\text{minimize}} \quad
& & \sum_{t=0}^{N} \left(\mathbf{x}_{ref,t} - \mathbf{x}_{t}\right)Q\left(\mathbf{x}_{ref,t} - \mathbf{x}_{t}\right) + \mathbf{u}_{t}R\mathbf{u}_{t} \\
& \textrm{subject to} \quad
%& & f_i(x) \leq b_i, \; i = 1, \ldots, m.\\
& & \mathbf{x}_0 = \mathbf{x}_t,\\
& & & \mathbf{x}_{k+1} = A\mathbf{x}_{k} + B\mathbf{u}_{k},\\
& & & ||\mathbf{v}_{k}|| \le v_{max}, \ ||\mathbf{a}_{k}|| \le a_{max}\\
& & & \text{P}\left({\color{black}{\mathbf{m}^T {\mathbf{v}^{orca}_{i}-b\ge0}}}\right) \ge \delta,\\
& & & \forall k=0, 1, ..., N-1.
\end{aligned}
\label{eqn:optimization}
\end{equation}
% \begin{equation}
% \begin{aligned}
% & \underset{}{\text{minimize}} \quad
% & & \sum_{t=0}^{N} \left(\pmb{\zeta}_{ref,t} - \pmb{\zeta}_{t}\right)Q\left(\pmb{\zeta}_{ref,t} - \bm{\zeta}_{t}\right) + \bm{\nu}_{t}R\pmb{\nu}_{t} \\
% & \textrm{subject to} \quad
% %& & f_i(x) \leq b_i, \; i = 1, \ldots, m.\\
% & & \pmb{\xi}_0 = \bm{\xi}_t,\\
% & & & \bm{\xi}_{k+1} = A\bm{\xi}_{k} + B\bm{\nu}_{k},\\
% & & & |v_{x,k}| \le V_{max}, |v_{y,k}| \le V_{max}, |v_{z,k}| \le V_{max},\\
% & & & |a_{x,k}| \le A_{max}, |a_{y,k}| \le A_{max}, |a_{z,k}| \le A_{max},\\
% & & & \text{P}\left(\mathbf{a}^T {v_{rvo}^{i}-b\ge0}\right) \ge \delta,\\
% & & & \forall k=0, 1, ..., N-1.
% \end{aligned}
% \label{eqn:optimization}
% \end{equation}
{\color{black} Our MPC-based algorithm} plans in terms of acceleration and yaw rate, {\color{black}which} constitute the control input (Section~\ref{sec:DCAD}). The constraint $\text{P}\left({\color{black}{\mathbf{m}^T {\mathbf{v}^{orca}_{i}-b\le0}}}\right) \le \delta$ {\color{black}represents} the chance constraints defined on ORCA, i.e. the constraint is said to be satisfied if{\color{black},} given the uncertainty in state, the probability that {\color{black}the} ORCA constraint is satisfied is greater than $\delta$. The output of the MPC is the control input vector for the quadrotor.

\subsection{Collision Avoidance Velocity}
The VO is constructed using the relative position ({\color{black}$\mathbf{r}_i-\mathbf{r}_j$}) and velocity ({\color{black}$\mathbf{v}_A-\mathbf{v}_B$}) of the agents. From {\color{black}Fig.~\ref{fig:Orca_traj}}, we notice that the ORCA {\color{black}plane} passes through the origin. Thus{\color{black},} the parameter $\color{black}{b}$ in the constraint~({\color{black}\ref{forcaE}}) is zero in this case. The feasible velocity set for agent A is constructed by translating this plane by {\color{black}the} agent's mean velocity plus $0.5$ times ({\color{black}}as in~\cite{ORCA}) the change in relative velocity proposed by ORCA. {\color{black}Due to this translation, $b$ can be non-zero,} and we use a mean value for $\color{black}{b}$ instead of a distribution to apply the formulation below.
%The resulting feasible set of velocities is constituted by the relative velocities between the two agents that are collision-free. To compute a collision-free velocity for agent A, the agent's mean velocity is translated by $0.5$ times the change in relative velocity proposed by chance-constrained ORCA. 

\subsection{Chance Constraint Formulation}\label{subsec:ccf}
% \begin{figure}
%     \centering
%     \includegraphics[width=2.4in, height=2.2in]{ieeeconf-3/Screen Shot 2020-02-27 at 2.25.07 AM.png}
%     \caption{Two robot Scenario}
%     \label{fig:my_label}
% \end{figure}
% \begin{figure}
%     \centering
%     \includegraphics[width=2.4in, height=2.2in]{ieeeconf-3/Screen Shot 2020-02-27 at 12.44.34 PM.png} 
%     \caption{Deterministic ORCA}
%     \label{fig:my_label}
% \end{figure}
% \begin{figure}
%     \centering
%     \includegraphics[width=2.4in, height=2.2in]{ieeeconf-3/Screen Shot 2020-02-27 at 1.39.42 PM.png} 
%     \caption{Chance Constraint}
%     \label{fig:my_label}
% \end{figure}
%Screen Shot 2020-02-27 at 2.25.07 AM
Since the position and velocity data of the agent and its neighbors are non-Gaussian random variables, the collision avoidance constraints {\color{black}must} consider the uncertainty in the agent's state estimations for safety. As mentioned in Section III-A, we do not make any assumption on the nature of the uncertainty distribution. However, we model uncertainty using two different methods. Our first method approximates the noise distribution as a Gaussian distribution, which works well for certain sensors. In comparison, our second method is more general and works with non-Gaussian uncertainty by fitting a Gaussian Mixture Model to the uncertainty distribution.\\
\textbf{Method I: Gaussian Distribution}\\
In this method, we approximate the position and velocity uncertainties using a multivariate Gaussian distribution. For an agent $i$, its position and velocity variables are approximated as
$\mathbf{r}_i \sim N(\boldsymbol{\mu}_{i,r}, {\Sigma_{i,r}})$, and $\mathbf{v}_i \sim N(\boldsymbol{\mu}_{i,v}, {\Sigma_{i,v}}).$
% %\[
%   \mathbf{r}_i \sim N(\boldsymbol{\mu}_{i,r}, {\Sigma_{i,r}}); \ \ \ \ \ \ \ \ \ \ \    \mathbf{v}_i \sim N(\boldsymbol{\mu}_{i,v}, {\Sigma_{i,v}}).
% %\]
The deterministic ORCA constraint between two agents $\color{black}{i}$ and $\color{black}{j}$ is given by the following plane (linear) equation{\color{black}:}
\begin{equation}
    f^{ORCA_{j}^{i}} (.) = {\color{black}{\mathbf{m}^T\mathbf{v}^{orca}_{i}-b}}.
    \label{forca}
\end{equation}
Here, the parameters $\color{black}{\mathbf{m}}$ and $\color{black}{b}$ are functions of the agent's trajectory. In a stochastic setting, the parameters $\color{black}{\mathbf{m}}$ and $\color{black}{b}$ are random variables due to their dependence on the agent's position and velocity. Though the uncertainty in position and velocity are assumed to be Gaussian for this case, this need not translate to a Gaussian distribution for $\color{black}{\mathbf{m}}$. {\color{black}However, we} approximate $\color{black}{\mathbf{m}}$'s distribution as Gaussian for the application of our {\color{black}algorithm}, with expectation and covariance represented by $\color{black}{\boldsymbol{\mu}_\mathbf{m}}$ and $\color{black}{\Sigma_\mathbf{m}}$, respectively. Eqn.~(\ref{eqn:chanceORCA}) {\color{black}highlights} the chance constraint, representing the probability that the ORCA constraint~(\ref{forca}) is satisfied, given the uncertainties in {\color{black}the} position and velocity data. This probability is set to be above a predefined confidence level ($\delta$). 
\begin{align}\label{eqn:chanceORCA}
  \begin{split}
     P({\color{black}{\mathbf{m}^T\mathbf{v}^{orca}_{i}-b \ge 0}}) \ge \delta, \ \
     {\color{black}{\mathbf{m} \sim N(\boldsymbol{\mu}_\mathbf{m}, \Sigma_\mathbf{m}). }}    
  \end{split}
\end{align}
From \cite{Prekopa}, we know that if $\mathbf{a}$ follows a Gaussian distribution, the chance constraint can be transformed into a deterministic second-order cone constraint. This is summarized in Lemma \ref{lemma:1}. %For the sake of clarity we use the same symbols as in \cite{Zhu,Blackmore}.
\begin{lemma}\label{lemma:1}
For a multivariate random variable $\color{black}{\mathbf{m} \sim N(\boldsymbol{\mu}_\mathbf{m}, \Sigma_\mathbf{m})}$, the chance constraint $\color{black}{P(\mathbf{m}^T\mathbf{x_t} \le b) > \delta}$  can be reformulated as a deterministic constraint on the mean and covariance.
{\small
\begin{multline}
    \color{black}{P(\mathbf{m}^T\mathbf{x}_t \le b) > \delta 
    \iff b - \boldsymbol{\mu}_\mathbf{m}^T \mathbf{x}_t \ge \textnormal{erf}^{-1}(\delta)        \left\Vert{{\Sigma}_\mathbf{m}^{\frac{1}{2}}\mathbf{x}_t}\right\Vert_2},
\end{multline}
}
where \textnormal{erf}(x) is the standard error function given by,
$\textnormal{erf}(x) = \frac{1}{2\pi}\int_{0}^{x}e^{-\tau^2/2}d\tau.$
\end{lemma}
Here, $\delta$ is the confidence level associated with satisfying the constraint $\color{black}{\mathbf{m}^T\mathbf{x_t} \le b}$.
Since our collision avoidance constraints are linear, each chance constraint can be {\color{black}reformulated} to a second order cone constraint{\color{black}, as shown} in Lemma~\ref{lemma:1}. Hence, each collision avoidance chance constraint can be written as
\begin{multline}
      {\color{black}{P(\mathbf{m}^T\mathbf{v}_{i}^{orca}-b \ge 0) \ge \delta}} \iff \\
      {\color{black}{\boldsymbol{\mu}_\mathbf{m}^T \mathbf{v}_{i}^{orca} - b + \textnormal{erf}^{-1}(\delta)        \left\Vert{{\Sigma}_\mathbf{m}^{\frac{1}{2}}\mathbf{v}_i^{orca}}\right\Vert_2 \le 0.}}
\end{multline}
%The distribution of `a' is approximated using a Gaussian model, with expectation and variance represented by $\mu_a$ and $\sigma_a^2$ respectively. %using the law of the unconscious statistician (LOTUS).
%For the function $g(x,y)$ of two random variables $x$ and $y$, the expectation is given by Eqn. \ref{eqn:LOTUS}.
%\begin{equation}\label{eqn:LOTUS}
%    E[g(x,y)] = \int_{-\infty}^{\infty}\int_{-\infty}^{\infty} g(x,y)f(x,y)dx dy
%\end{equation}
%The function $f(x,y)$ is the joint probability density.
%In our case, the random multivariate function is `a'.\\\\ 
% \begin{equation}\label{eqn:LOTUS}
%     \begin{split}
%     %E[a(\mathbf{r}^i,\mathbf{r}^j, \mathbf{v}^i, \mathbf{v}^j)] 
%     \mu_a=
%         \int_{-\infty}^{\infty}...\int_{-\infty}^{\infty} a(.)f(.)dr_x^i dr_y^i dr_z^i dv_x^i dv_y^i dv_z^i\\
%     dr_x^j dr_y^j dr_z^j dv_x^j dv_y^j dv_z^j.
%     \end{split}
% \end{equation}
% \begin{equation}
%     (\sigma_a)^2 = E[(a - E[a])^2]
% \end{equation}
\textbf{Method II: Gaussian Mixture Model (GMM)}\\
To handle non-Gaussian uncertainty distributions, we present an extension of the Gaussian formulation (Method I). {\color{black}In this case}, the probability distribution for the position and velocity is assumed to be non-parametric and non-Gaussian, i.e. the probability distribution {\color{black}{is}} not known. We assume that we have access to ${\color{black}s}$ samples of these states that could come from a black-box simulator or a particle filter. Using ${\color{black}s}$ samples, a distribution for the parameter $\color{black}{\mathbf{m}}$ is constructed. In our implementation{\color{black}, we use a value of} ${\color{black}s}=40$ samples.%For simplicity, we approximate this uncertainty distribution using a GMM. 

A GMM model of ${\color{black}n}$ Gaussian components is fit to the probability distribution of parameter $\color{black}{\mathbf{m}}$ in Eqn.~\ref{forca} using Expectation-Maximization (EM). Each collision avoidance constraint is split into ${\color{black}n}$ second-order constraints, each corresponding to a single Gaussian component. Furthermore, an additional constraint is {\color{black}used} that relates the $\mathbf{n}$ second-order constraints to GMM probability distribution. From~\cite{hu2018chance}, we know that if $\color{black}{\mathbf{m}}$ follows a GMM distribution, the chance constraint can be transformed into a set of {\color{black}deterministic} constraints. This is summarized in Lemma \ref{lemma:2}.

\begin{lemma}\label{lemma:2}
For a non-Gaussian random variable $\color{black}{\mathbf{m}}$ and a linear equation $f(\mathbf{x_t}) = \color{black}{\mathbf{m}^T\mathbf{x_t} \le b}$, the chance constraint $P({\color{black}{\mathbf{m}^T\mathbf{x_t} \le b}}) > \delta$ can be reformulated as a set of deterministic constraints on the mean and co-variance.
Let the distribution of $\color{black}{\mathbf{m}}$ be approximated by $\mathbf{n}$ Gaussian components. The probability that $f(\mathbf{x_t})$ is satisfied while $\color{black}{\mathbf{m}}$'s distribution is given by the $n^{th}$ Gaussian component of the GMM is given by
\begin{equation}
    %P_{i} = \textnormal{erf}\bigg(\frac{b - \boldsymbol{\mu}_{a}^T\mathbf{x}_t}{\sqrt{\mathbf{a}^T\Sigma_t \mathbf{a}}}\bigg).
    {\color{black}{P_{i} = \textnormal{erf}\bigg(\frac{b - \boldsymbol{\mu}_{\mathbf{m}}^T\mathbf{x}_t}{\sqrt{\mathbf{x_t}^T\Sigma_\mathbf{m} \mathbf{x_t}}}\bigg).}}
\end{equation}
Then, the probability that $f(\mathbf{x_t})$ is satisfied for GMM distribution of $\color{black}{\mathbf{m}}$ is given by
\begin{equation}
    P_{GMM} =\sum_{i=1}^{n} {\color{black}\alpha_iP_{i}}.
\end{equation}
Here, ${\color{black}\alpha_i}$s are the mixing coefficients for the GMM, satisfying $\sum_{i=1}^n {\color{black}\alpha_{i}} = 1$.
\end{lemma}
Let us assume that the probability of satisfying $\color{black}{\mathbf{m}^T\mathbf{v}_{i}^{orca}-b \ge 0,}$ while considering only the $n^{th}$ Gaussian component{\color{black},} is given by $\eta_{i}$. We can reformulate the constraint using Lemma~\ref{lemma:2}, which is given by Eqn.~\ref{eqn:GMM1}.

% Considering a Gaussian component `i' of the GMM, the probability that Eqn.~\ref{forca} is satisfied can be given as:
% \begin{equation}
%     P_{i} = \textnormal{erf}\bigg(\frac{(b - \bar{a}^T)}{\sqrt{a^T\Sigma_t a}}\bigg).
% \end{equation}
% Moreover, the probability distribution function for the GMM is given by,
% \begin{equation}
%     P_{GMM} =\sum_{i=1}^{n} \phi_i*P_{i}
% \end{equation}
% Here, $\phi_i$'s are the mixing coefficients satisfying $\sum_{i=1}^n \phi_{i} = 1$.
% Lets assume that the confidence of satisfying $a\mathbf{v}_{rvo}^{i}-b \ge 0$, while considering the $i^{th}$ Gaussian component is given by $\eta_{i}$. We can reformulate the constraint using Lemma~\ref{lemma:1}.
% \begin{equation}
%     \bar{a}^T {v_{rvo}} - b + \textnormal{erf}^{-1}(\eta_i){\left\Vert{\Sigma_a^{1/2} v_{rvo}}\right\Vert} \le 0, i \in \{1,2,...,n\}.
% \end{equation}
The probability of satisfying the linear constraint considering the GMM distribution for $\color{black}{\mathbf{m}}$ is computed by mixing the individual component probabilities using the mixing coefficients ({\color{black}$\alpha_i$}). A probability $\delta$ is chosen as the required confidence level.
Eqn.~(\ref{eqn:GMMCC}) represents the chance constraint that the probability of satisfying {\color{black}Eqn.~(\ref{forca})} is greater than $\delta$. The chance constraint can be reformulated as:
\begin{equation*}
    \resizebox{0.5\linewidth}{!}{ $P({\color{black}{\mathbf{m}^T\mathbf{v}_{i}^{orca}-b \ge 0}}) \ge \delta \iff $}
\end{equation*}
%\begin{align}
\begin{numcases} {} 
   \resizebox{0.85\linewidth}{!}{ $
   \begin{aligned}\label{eqn:GMM1}
        {\color{black}{\boldsymbol{\mu}_{\mathbf{m}}^T {\mathbf{v}_{i}^{orca}} - b + \textnormal{erf}^{-1}(\eta_i){\left\Vert{\Sigma_\mathbf{m}^{1/2} \mathbf{v}_{i}^{orca}}\right\Vert}_2}} \le 0,  \forall i \in \text{1 to n }
    \end{aligned}
    $}\\
    \resizebox{0.25\linewidth}{!}{$\sum_{i=1}^n {\color{black}{\alpha_i\eta_i > \delta}}$}. \label{eqn:GMMCC} 
\end{numcases}
%\end{align}
% \begin{equation}
%     \sum_{i=1}^n \phi_i*\eta_i > \delta,
% \end{equation}
In Eqn.~\ref{eqn:GMMCC} the values for the mixing coefficient and confidence are known prior to the optimization. We notice that for a given set of mixing coefficients ({\color{black}$\alpha_i$s}) and confidence ($\delta$), multiple sets of values for {\color{black}$\eta_i$'s} can satisfy Eqn.~\ref{eqn:GMMCC}. The value of  $\eta_i$ in turn affects the feasible velocity set. Hence, we plan for  {\color{black}$\eta_i$'s} in addition to the control input in problem~({\color{black}Equation}~\ref{eqn:optimization}). When GMM has 3 components, i.e. ${n} = 3$, we have three additional variables given by $\eta_1, \eta_2, \eta_3$ in the optimization problem.
%$$\mathbf{u} = [a_x, a_y, a_z, \dot{\psi}, \eta_1, \eta_2, \eta_3].$$
Now the optimization problem~({\color{black}Equation}~\ref{eqn:optimization}) simultaneously computes values for acceleration, $\dot{\psi}$ and $\eta_i${\color{black},} such that the collision avoidance chance constraints (Eqn.~\ref{eqn:GMMCC}) are satisfied.

The computed control input from the optimization problem~(\ref{eqn:optimization}) is transformed into an inner-loop control input for the quadrotor using an inverse map{\color{black},} similar to~\cite{DCAD}.
\section{Results}
\begin{figure*}[t]
  \centering
  \begin{subfigure}[b]{0.32\textwidth}
  \includegraphics[height=4.0cm,width=1\linewidth]{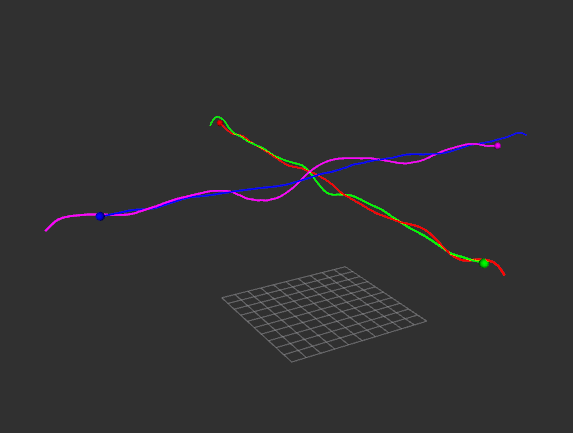}
  \caption{DCAD(Deterministic)}
  \label{fig:proposedMethod_Vel}
  \end{subfigure}
  \begin{subfigure}[b]{0.32\textwidth}
  \includegraphics[height=4.0cm,width=1\linewidth]{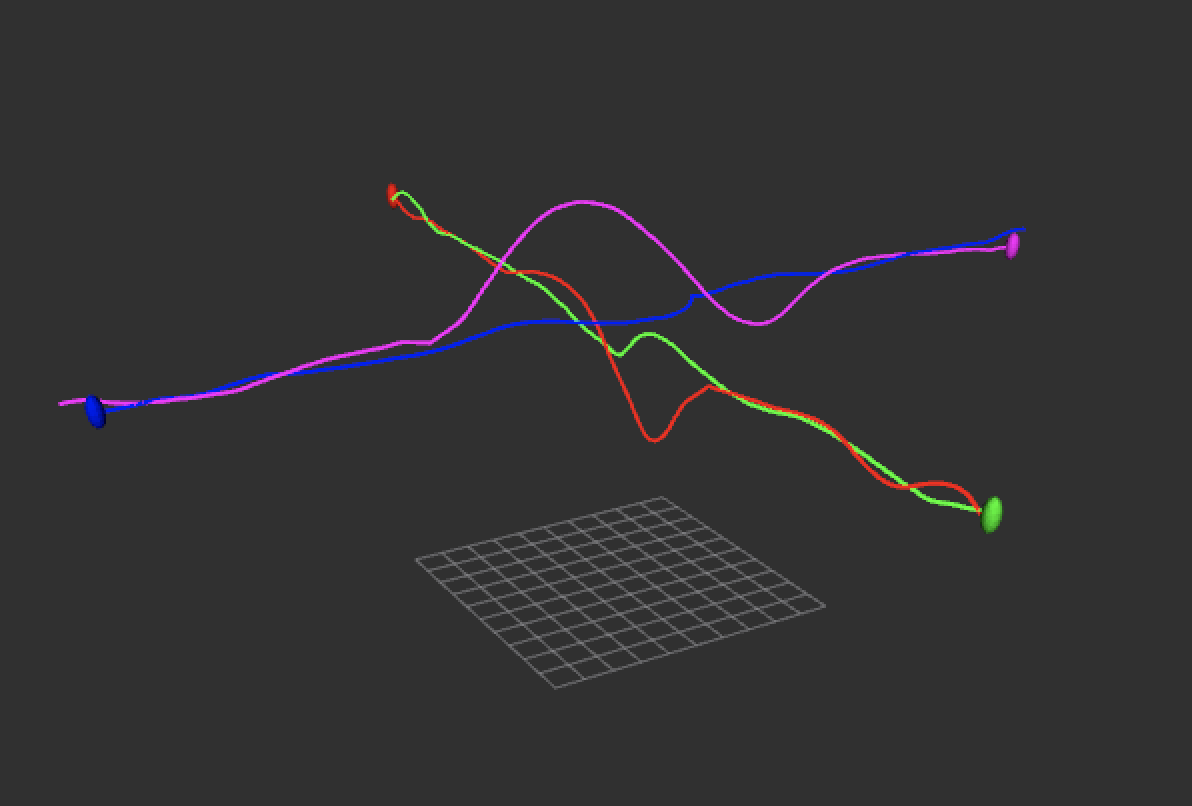}
  \caption{Gaussian Noise Approximation}
  \label{fig:ORCA_Vel}
  \end{subfigure}
  \begin{subfigure}[b]{0.32\textwidth}
  \includegraphics[height=4.0cm,width=1\linewidth]{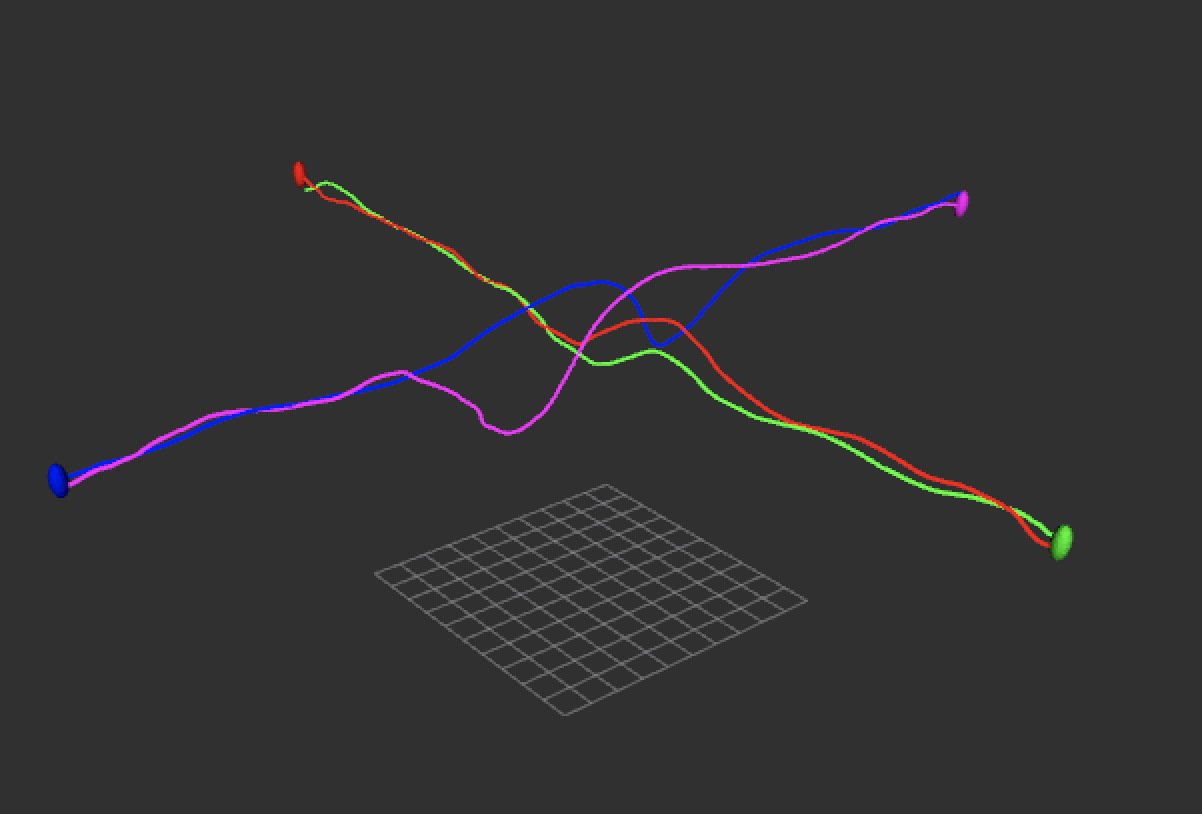}
  \caption{Non-Gaussian Noise Approximation}
  \label{fig:AVO_Vel}
  \end{subfigure}
      \caption{The figure illustrates a scenario with 4 agents exchanging positions with their diagonally opposite {\color{black}agents}. The collision avoidance is performed based on deterministic DCAD and Gaussian and non-Gaussian SwarmCCO. {\color{black}{DCAD results in more collisions because uncertainty is not explicitly considered}}. We observe that {\color{black}some agents travel} a {\color{black}longer} path when using the Gaussian formulation {\color{black}{than}} when using the non-Gaussian formulation (Section~\ref{GvsNG:PL}).%The figure illustrates a scenario with 4 agents exchanging positions with their diagonally opposite agent. The collision avoidance is performed based on DCAD and Gaussian and non-Gaussian SwarmCCO. DCAD results in collisions as uncertainty is not explicitly considered, in this figure the agent grazed past each other. We observe that the agent travels a larger path when using the Gaussian formulation that when using the non-Gaussian formulation (Section~\ref{GvsNG:PL}). That is, the Gaussian method is relatively conservative. The trajectories for the Gaussian and non-Gaussian methods are from a trial where no collisions were observed.
      }
      \label{fig:velvar}
\end{figure*}
In this section, we describe the implementation of {\color{black}our algorithm} and our simulation setup. Further, we summarize our evaluations and {\color{black}highlight the benefits of our approach}.
%In this section, we describe the implementation of the proposed method and our simulation setup. Further, we summarize our evaluations and present our method's benefits.
\subsection{Experimental Setup}
Our method is implemented on an Inter Xeon w-2123 \@ 3.6 GHz with 32 GB RAM and a GeForce GTX 1080 GPU. Our simulations are built using the PX4 Software In The Loop (SITL) framework, ROS Kinetic, and Gazebo 7.14.0. We solve the MPC optimization using the IPOPT Library with a planning horizon of $N = 8$ steps and a time step of $\delta$t = $100$ms. We {\color{black}consider} a non-Gaussian distribution for the position and velocity data, {\color{black}from which the input sensor readings for both the Gaussian and non-Gaussian SwarmCCO methods are generated}. Gazebo state information represents the ground truth position and velocity data, while we add non-zero mean, non-Gaussian noise to simulate state uncertainties. The added noise is generated through a GMM model of 3 Gaussian components. We use two different GMM to simulate noise for position readings{\color{black}:} $GMM_1:  \mu_{GMM1} = [0.15, 0.08, -0.05]$, {\color{black}$\Sigma_{GMM1} = diag([0.06, 0.7, 0.3])$,} and $GMM_2: \mu_{GMM2} = [0.2, 0.0, -0.2]$, {\color{black}$\Sigma_{GMM2} = diag([1.0, 0.3, 1.0])$.}  The velocity readings are simulated using a noise distribution that {\color{black}has} half the mean and covariance values of $\Sigma_{GMM1}$ and $\Sigma_{GMM2}$. Further, for our evaluation we consider two confidence levels given by $\delta_1 = 0.75$ and $\delta_2 = 0.90$.  %We fit the GMM model for parameter $a$ using the Armadillo library {\color{red} [cite]}. 
The RVO-3D library is utilized to compute the ORCA collision avoidance constraints. We consider a sensing region of 8m for {\color{black}the} ORCA plane computation. {\color{black}{The agent's physical radius is assumed to be $0.25m$ while the agent's radius for the ORCA computation is set as $0.5m$.}} In our evaluations, two agents are considered to be in collision if their positions are less than $0.5m$ apart. Further, we show results for {\color{black}the} non-Gaussian method with 2 {\color{black}components} ($n=2$) GMM and 3 {\color{black}components} ($n=3$) GMM.

\subsection{Generated Trajectories}
We evaluate our method in simulation with four quadrotors exchanging positions with the antipodal agents (circular scenario). Fig.~{{\ref{fig:velvar}}} shows the resulting trajectories for this scenario using deterministic DCAD~\cite{DCAD} and Gaussian and non-Gaussian SwarmCCO. We observe that in deterministic DCAD, the agents do not handle noise and hence they trajectories often result in collisions. In Fig.~\ref{fig:velvar} the DCAD trajectories are such that the agent graze past each other. In contrast, the trajectories generated by SwarmCCO methods are safer. %We observe that the trajectories generated using the Gaussian method are more conservative than those generated using the non-Gaussian method and result in large deviations of the agent from its straight-line reference path to the goal. 

\subsection{Collision Avoidance}
\begin{table*}
\caption{Number of episodes in which one or more quadrotor collided when the agents out of a total of 100 episodes. The DCAD algorithm is deterministic and does not consider the uncertainty in state. We observe that our Gaussian and non-Gaussian formulations provide improved safety.}\label{tab:Table 2}
\centering
\renewcommand{\arraystretch}{1.3}
\begin{tabular}{|c||c|c|c|c|c|c|c|}
\hline
\multirow{3}{*}{Noise} & \multirow{3}{*}{Confidence Level} & \multirow{3}{*}{Method} & \multicolumn{5}{c|}{Number of collisions}\\
\cline{4-8}
& & & \multicolumn{5}{c|}{No of Agents} \\
\cline{4-8}
& & & 2 & 4 & 6 & 8 & 10 \\
\hline
\multirow{8}{*}{$\Sigma_1$} & \multirow{4}{*}{$\delta=0.75$} 
& DCAD (deterministic) & 27 & 25 & 35 & 40 & 44  \\
& & Gaussian SwarmCCO & 0 & 4 & 4 & 7 & 15 \\
%& & Non-Gaussian SwarmCCO (n=2) & 0 & 3 & 9 & 11 & 24\\
& & Non-Gaussian SwarmCCO (n=2) & 0 & 2 & 5 & 4 & 12\\
& & Non-Gaussian SwarmCCO (n=3) & 0 & 0 & 2 & 5 & 13\\
\cline{2-8}                                                            
& \multirow{4}{*}{$\delta=0.90$}
& DCAD (deterministic) & 27 & 25 & 35 & 40 & 44 \\
& & Gaussian SwarmCCO & 0 & 3 & 2 & 7 & 13 \\
%& & Non-Gaussian SwarmCCO (n=2) & 0 & 5 & 5 & 7 & 9\\
& & Non-Gaussian SwarmCCO (n=2) & 0 & 1 & 3 & 4 & 12 \\
& & Non-Gaussian SwarmCCO (n=3) & 0 & 0 & 1 & 2 & 10 \\
\hline
\multirow{8}{*}{$\Sigma_2$} & \multirow{4}{*}{$\delta=0.75$} 
& DCAD (deterministic) & 56 & 26 & 32 & 51 & 58 \\
& & Gaussian SwarmCCO & 0 & 2 & 3 & 11 & 18   \\
%& & Non-Gaussian SwarmCCO (n=2) & 0 & 3 & 5 & 1 & 13 \\
& & Non-Gaussian SwarmCCO (n=2) & 0 & 1 & 3 & 5 & 9 \\
& & Non-Gaussian SwarmCCO (n=3) & 0 & 0 & 0 & 3 & 10 \\
\cline{2-8}                                                            
& \multirow{4}{*}{$\delta=0.90$}
& DCAD (deterministic) & 56 & 26 & 32 & 51 & 58  \\
& & Gaussian SwarmCCO & 0 & 2 & 1 & 4 & 6  \\
%& & Non-Gaussian SwarmCCO (n=2) & 0 & 0 & 2 & 4 & 8 \\
& & Non-Gaussian SwarmCCO (n=2) & 0 & 0 & 1 & 3 & 4  \\
& & Non-Gaussian SwarmCCO (n=3) & 0 & 0 & 0 & 2 & 4  \\
\hline
\end{tabular}
\end{table*}

{\color{black}{We evaluate our method in a circular scenario}}. Table~\ref{tab:Table 2} summarizes the number of trials with observed collisions out of a total of 100 trials. {\color{black}{We observe that the performance of the deterministic method degrades (in terms of number of collisions) with added noise}}.% It is clear that the deterministic method heavily deteriorates in performance with added noise.
{\color{black}{ In contrast, we}} observe good performance with both the Gaussian and non-Gaussian SwarmCCO. {\color{black}{As expected, the number of collisions reduces with}} an increase in confidence level ($\delta$). %This improvement is visible with increase in the number of agents in the environment.
%Table~\ref{tab:Table 2} summarizes the number of trials with observed collisions out of a total of 100 trials. It is clear that the deterministic method heavily deteriorates in performance with added noise. We observe good performance with both the Gaussian and non-Gaussian SwarmCCO. With an increase in confidence level ($\eta$), the number of collisions reduces further. This improvement is visible with increase in the number of agents in the environment.

\subsection{Gaussian vs. Non-Gaussian:}
In this subsection we compare the performance of our Gaussian and non-Gaussian formulations for SwarmCCO. We consider the circular scenario where the agents move to their antipodal positions.
\subsubsection{Path Length}\label{GvsNG:PL}
\begin{table*}
\caption{Average path length traveled by the agent while exchanging positions with the antipodal agents. The reference, straight line path to the goal is $40$m long. The Gaussian method is relatively conservative, resulting in longer path lengths on average compared to the non-Gaussian method. We observe this performance in scenarios with 8 and 10 agents.%The number of trials (out of 100) with quadrotor collisions when agents travel to their antipodal positions.  We observe that our algorithms result in fewer collisions, as compared to deterministic DCAD. We consider two cases for the non-Gaussian SwarmCCO, the first case with two Gaussian components (n=2), and the other with 3 components (n=3). %Average path length travelled by the agent when exchanging positions with the antipodal agents. Gaussian method is relatively conservative resulting in longer path lengths on average compared to the non-Gaussian method. This is especially visible as the number of agents in the environment increases.
}\label{tab:Table 3}
\centering
\renewcommand{\arraystretch}{1.3}
\begin{tabular}{|c||c|c|c|c|c|c|c|}
\hline
\multirow{3}{*}{Noise} & \multirow{3}{*}{Confidence Level} & \multirow{3}{*}{Method} & \multicolumn{5}{c|}{Path Length}\\
\cline{4-8}
& & & \multicolumn{5}{c|}{No of Agents} \\
\cline{4-8}
& & & 2 & 4 & 6 & 8 & 10 \\
\hline
\multirow{6}{*}{$\Sigma_1$} & \multirow{3}{*}{$\delta=0.75$} 
& Gaussian & 41.12 & 42.75 & 44.04 & 44.88 & 47.92  \\
%& & Non-Gaussian (n=2) & 41.03 & 42.14 & 43.40 & 44.49 & 46.49\\
& & Non-Gaussian (n=2) & 41.06 & 41.95 & 43.03 & 44.39 & 46.66 \\
& & Non-Gaussian (n=3) & 41.06 & 42.09 & 43.12 & 44.59 & 46.52 \\
\cline{2-8}                                                            
& \multirow{3}{*}{$\delta=0.90$}
& Gaussian & 41.10 & 43.14 & 45.05 & 46.64 & 48.87  \\
%& & Non-Gaussian (n=2) & 41.07 & 42.54 & 43.70 & 44.59 & 46.93\\
& & Non-Gaussian (n=2) & 41.07 & 42.03 & 43.31 & 44.42 & 46.53 \\
& & Non-Gaussian (n=3) & 41.06 & 42.12 & 43.35 & 44.57 & 46.62 \\
\hline
\multirow{6}{*}{$\Sigma_2$} & \multirow{3}{*}{$\delta=0.75$} 
& Gaussian & 41.21 & 43.06 & 44.51 & 46.38 & 49.61   \\
%& & Non-Gaussian (n=2) & 41.19 & 42.81 & 44.28 & 46.30 & 48.50 \\
& & Non-Gaussian (n=2) & 41.21 & 42.67 & 44.09 & 45.80 & 47.87  \\
& & Non-Gaussian (n=3) & 41.22 & 42.47 & 44.14 & 45.93 & 48.15 \\
\cline{2-8}                                                            
& \multirow{3}{*}{$\delta=0.90$}
& Gaussian & 41.23 & 43.69 & 45.37 & 47.90 & 50.41  \\
%& & Non-Gaussian (n=2) & 41.25 & 43.39 & 45.06 & 46.90 & 49.30 \\
& & Non-Gaussian (n=2) & 41.21 & 42.47 & 44.13 & 45.68 & 48.09  \\
& & Non-Gaussian (n=3) & 41.23 & 42.70 & 44.62 & 46.13 & 48.25 \\
\hline
\end{tabular}
\end{table*}

{\small
\begin{table*}
\caption{Mean time required by the agents to reach the goal. The mean time required is approximately the same for all the methods due to the trajectory tracking MPC uses in our formulation.}\label{tab:Table 4}
\centering
\renewcommand{\arraystretch}{1.3}
\resizebox{18cm}{!}{%
\begin{tabular}{|c||c|c|c|c|c|c|c|c|c|c|c|c|c|c|c|c|}
\hline
\multirow{3}{*}{Method} & \multicolumn{8}{c|}{$\Sigma_1$} & \multicolumn{8}{c|}{$\Sigma_2$}\\
\cline{2-17}
& \multicolumn{4}{c|}{$\delta=0.75$} & \multicolumn{4}{c|}{$\delta=0.90$} & \multicolumn{4}{c|}{$\delta=0.75$} & \multicolumn{4}{c|}{$\delta=0.90$} \\
\cline{2-17}
& 2 & 4 & 6 & 8 & 2 & 4 & 6 & 8 & 2 & 4 & 6 & 8 & 2 & 4 & 6 & 8\\
\hline
Gaussian     
& 31.41 & 31.41 & 31.41 & 31.42 & 31.41 & 31.42 & 31.41 & 31.41 & 31.41 & 31.41 & 31.42 & 31.43 & 31.41 & 31.41 & 31.41 & 31.42\\
Non-Gaussian (n=2) 
& 31.41 & 31.41 & 31.41 & 31.42 & 31.41 & 31.42 & 31.41 & 31.41 & 31.41 & 31.41 & 31.42 & 31.41 & 31.41 & 31.41 & 31.41 & 31.41\\Non-Gaussian (n=3)
& 31.41 & 31.41 & 31.41 & 31.42 & 31.41 & 31.42 & 31.41 & 31.41 & 31.41 & 31.41 & 31.42 & 31.41 & 31.41 & 31.41 & 31.41 & 31.42\\
\hline
\end{tabular}
}
\end{table*}
}
For each agent, the reference path to its goal is a straight-line path of 40m length directed to the diametrically opposite position. In Table~\ref{tab:Table 3}, we tabulate the mean path length for the agents as they reach their goal while avoiding collisions. To compute this mean, we utilize only the trials that were collision-free. The mean is computed over 100 trials. We observe that the Gaussian method is (relatively) more conservative than the non-Gaussian method resulting in a longer path length for most scenarios.
\subsubsection{Time to Goal}
We observe that on average, the agents in all the methods reach their goal in the same time, this can be observed from Table~\ref{tab:Table 4}. This is due to the trajectory tracking MPC used by the agents, which modifies the agent velocity such that the agents reach their goal in approximately the same time duration ($\sim30s$).
\subsubsection{Inter-agent Distance}
\begin{figure*}
  \centering
  %\raisebox{25pt}{\parbox[b]{.1\textwidth}{4 Agents}}%
  \begin{subfigure}[b]{1\textwidth}
  {\includegraphics[height=3.8cm,width=.32\textwidth]{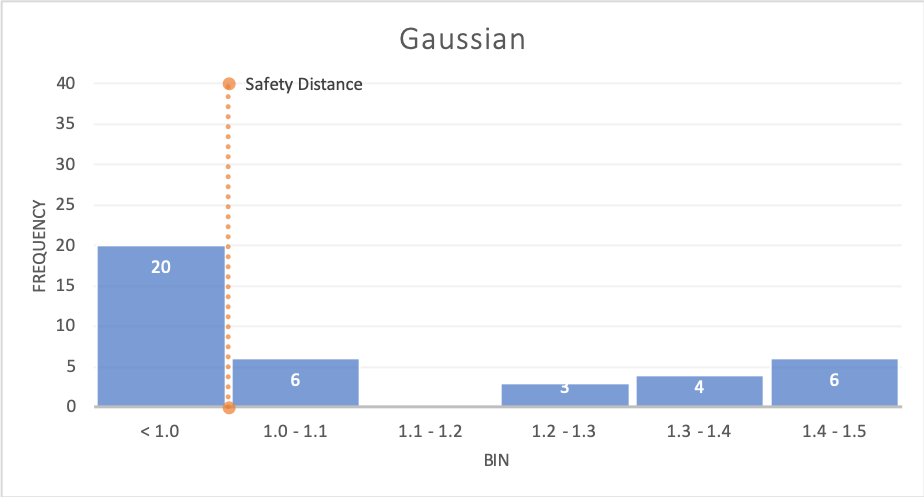}}\hfill
  {\includegraphics[height=3.8cm,width=.32\textwidth]{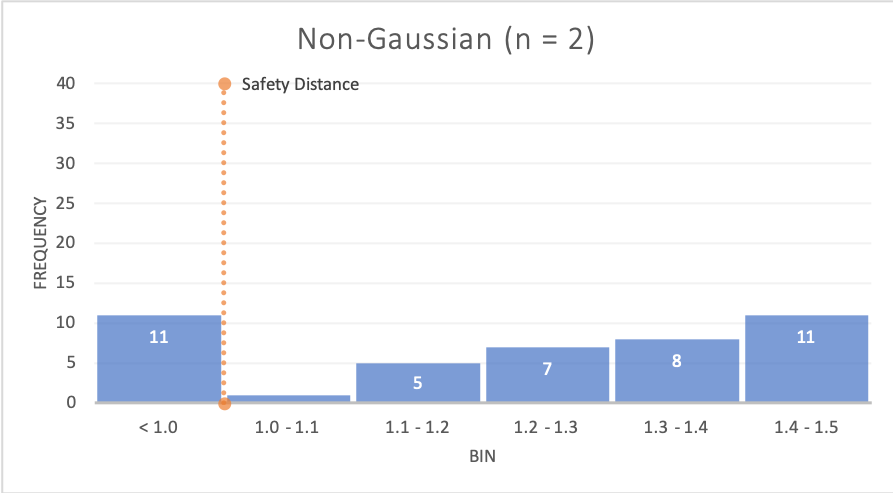}}\hfill
 {\includegraphics[height=3.8cm,width=.32\textwidth]{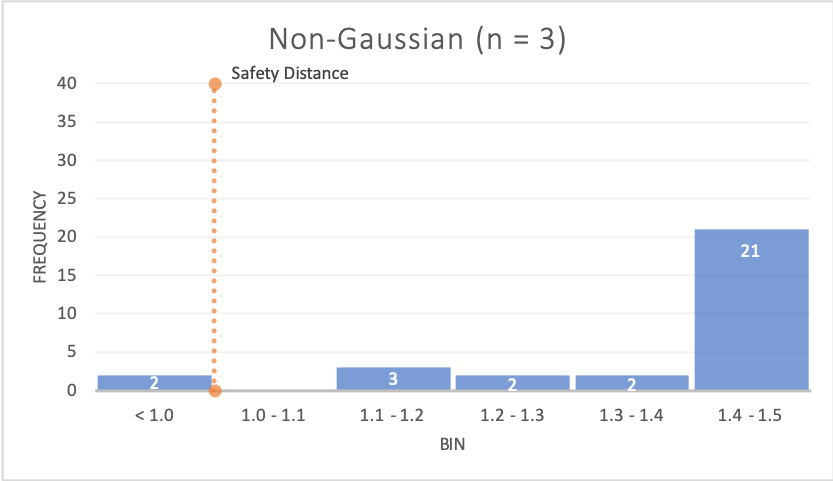}}\par
  \caption{Scenario with four agents.}
  %\vspace{2pt}
  \hfill
  \end{subfigure}
  %\raisebox{35pt}{\parbox[b]{.1\textwidth}{6 Agents}}%
  \begin{subfigure}[b]{1\textwidth}
  {\includegraphics[height=3.8cm,width=.32\textwidth]{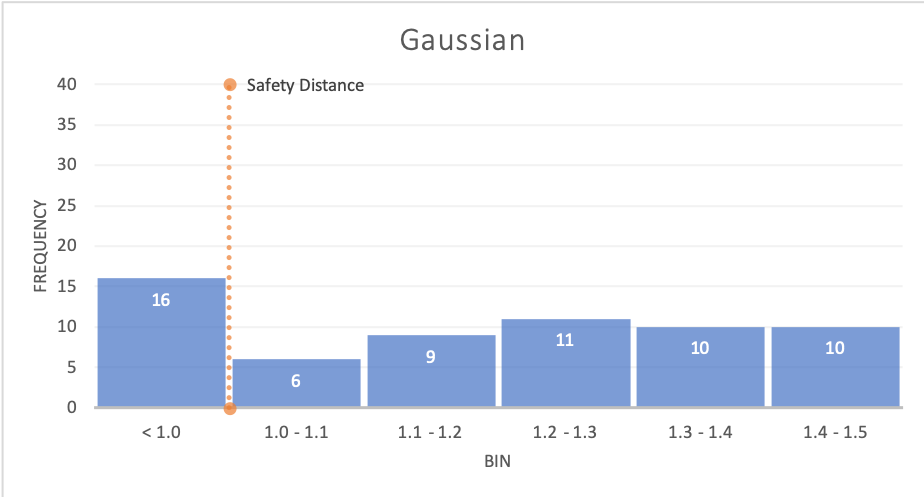}}\hfill
  {\includegraphics[height=3.8cm,width=.32\textwidth]{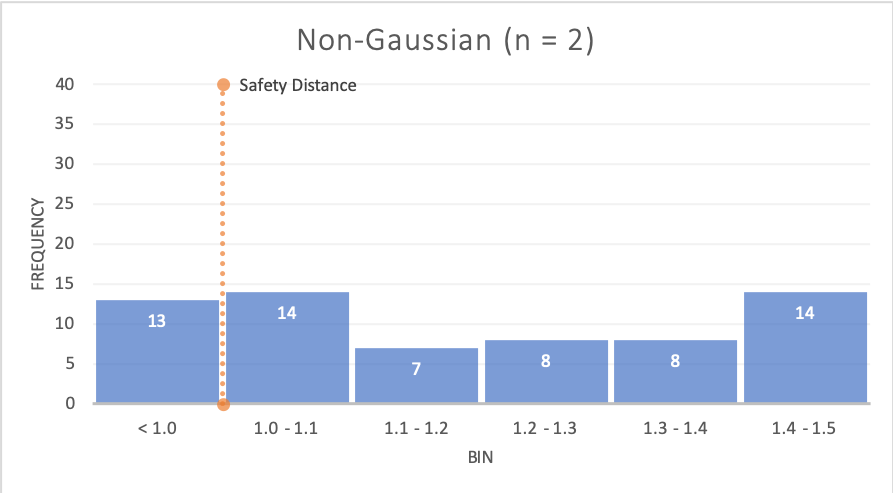}}\hfill
  {\includegraphics[height=3.8cm,width=.32\textwidth]{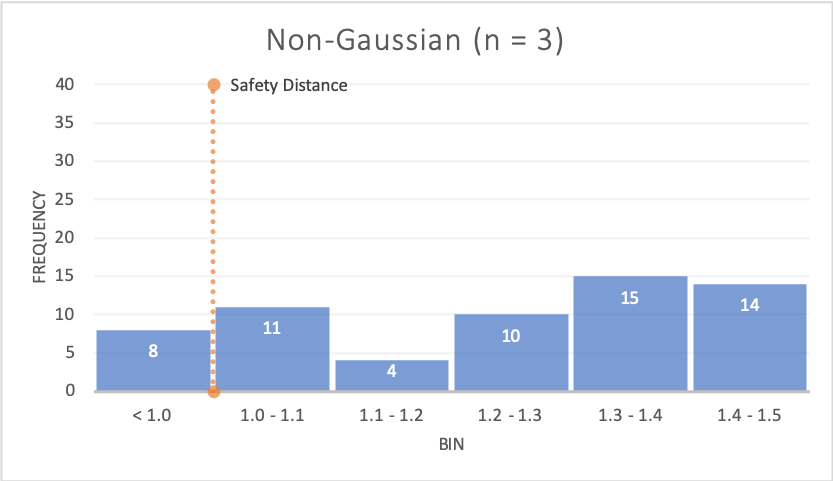}}\par
  \caption{Scenario with six agents.}
  %\vspace{2pt}
  \hfill
  \end{subfigure}
  %\raisebox{35pt}{\parbox[b]{.1\textwidth}{8 Agents}}%
  \begin{subfigure}[b]{1\textwidth}
  {\includegraphics[height=3.8cm,width=.32\textwidth]{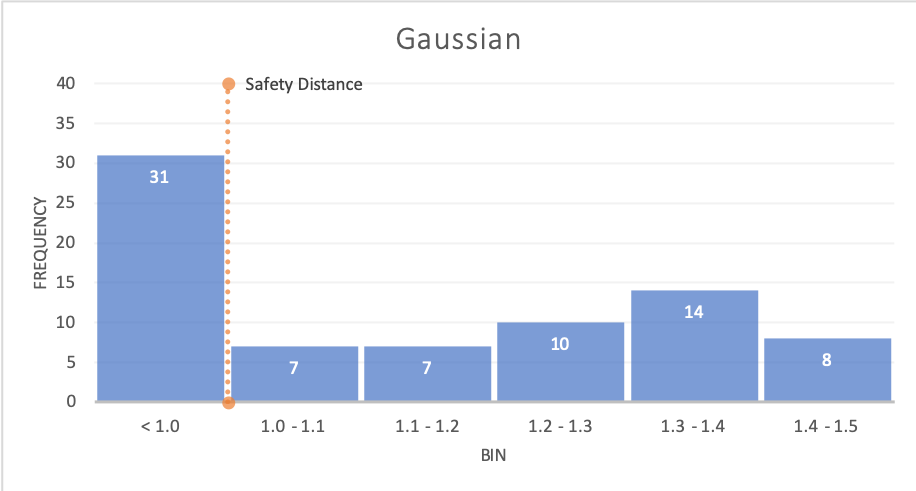}}\hfill
  {\includegraphics[height=3.8cm,width=.32\textwidth]{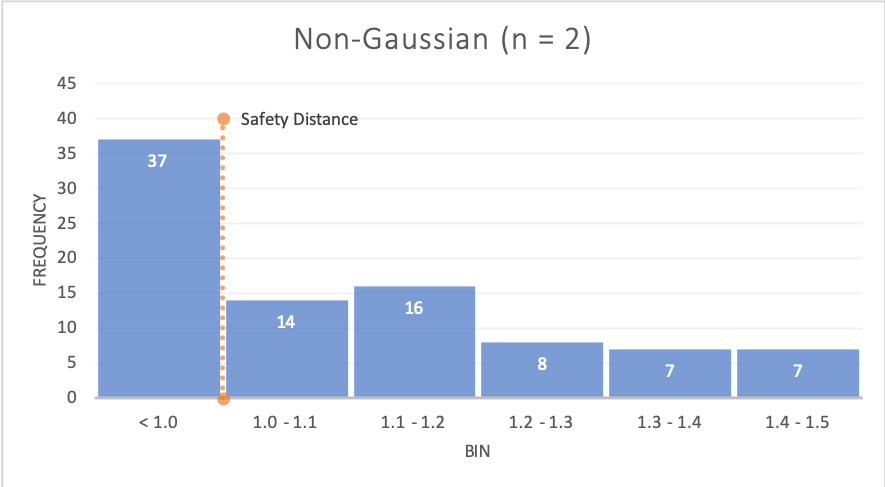}}\hfill
  {\includegraphics[height=3.8cm,width=.32\textwidth]{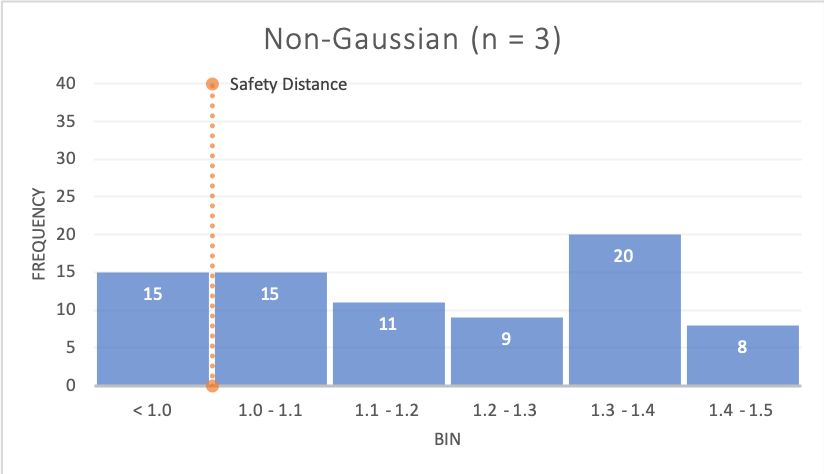}}
  \caption{Scenario with eight agents.}
  \hfill
  \end{subfigure}  
  \caption{Histogram of least inter-agent distance in the circular scenario with 4, 6 and 8 agents. The agents have a radius of $0.25m$, and the ORCA planes are constructed with an augmented agent radius of $0.5m$. Hence, an inter-agent distance of less than $1m$ is a collision according to the ORCA constraint, though the agents do not actually collide. The histogram is constructed over 100 trials, and the trials with least inter-agent distance greater than $1.5m$ are not included in the histogram. The safe inter-agent distance of $1m$ is denoted by the dotted orange line. We observe that the non-Gaussian method results in fewer trials with least inter-agent distance below $1m$, especially for the non-Gaussian formulation with 3 components.%Figure shows the histogram of the least inter-agent distance in a circular scenario with 4 agents. The agents have a radius of $0.25m$ and the ORCA planes are constructed with augmented agent radius of $0.5m$. Hence, an inter-agent distance of less than $1m$ is a collision according to the ORCA constraint though the agents do not actually collide. The histogram is constructed over a 100 trials and the number of trials with inter-agent distance greater than $1.5m$ is not included in the histogram. The safe inter-agent distance of $1m$ is dentoed by the dotted orange line in the figure. We observe that non-Gaussian method addresses the noise better resulting in lower number of trials with inter-agent distance below $1m$. This is especially the case with the non-Gaussian formulation with 3 Gaussian components. This is expected due to the (relatively) better modeling of the true noise distribution in the case of non-Gaussian method as compared to the Gaussian method.
  }
  \label{fig:ID}
\end{figure*}
In the ORCA computation, the agent radius is augmented to be $0.5m$ in contrast to the original agent radius of $0.25m$ to provide a safe distance around the agent. Thus, the safe inter-agent distance is $1m$. We compare the Gaussian and non-Gaussian formulations for the number of trials in which the safety threshold distance was compromised (out of 100 trials). We observe that the non-Gaussian method performs better in this case, and the results are illustrated through a histogram in Fig.~\ref{fig:ID}. We observe that with an increase in the number of agents in the environment, the inter-agent distance dips below $1m$ in multiple trials, but the non-Gaussian method with 3 Gaussian components performs better resulting in lower number of such trials. 

\subsection{Scalability}
\begin{figure}[t]
    \centering
    %\framebox{\parbox{3in}{We suggest that you use a text box to insert a graphic (which is ideally a 300 dpi TIFF or EPS file, with all fonts embedded) because, in an document, this method is somewhat more stable than directly inserting a picture.}}
    \includegraphics[width=0.99\linewidth]{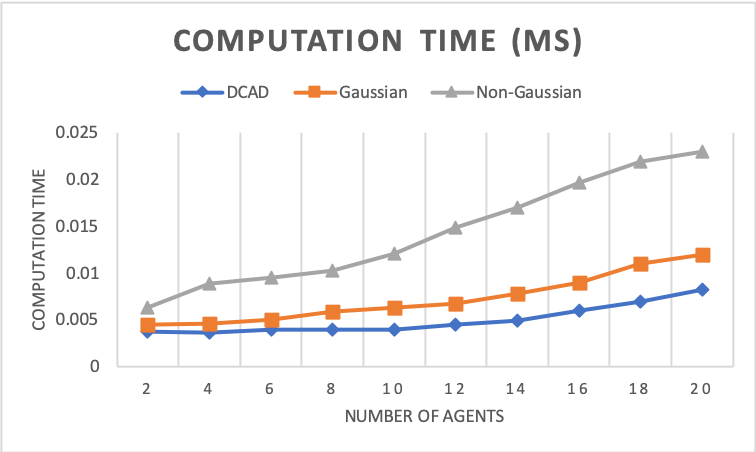}
    \caption{Control input computation time (ms) for one agent in the presence of 2 to 20 neighboring agents.}
    \label{fig:scalability}
\end{figure}

Figure~\ref{fig:scalability} illustrates the computation time (in milliseconds) of our algorithm for one agent with $1$ to $20$ neighbouring agents in the environment. From our previous work~\cite{DCAD}, and from our experiments we observe that considering the closest 10 obstacles provides good performance in most cases. We observe that, on average, our Gaussian method requires $\sim5ms$ to compute a collision avoiding input, while our non-Gaussian method requires $\sim7ms$ in the presence of $4$ neighbors.

%Considering ‘N’ agents in the environment and each agent considers ‘p’ neighbors in its sensing region. This gives a maximum of `P' linear collision avoidance constraints in addition to the constraints on the agent dynamics, state varaibles, and control input. Hence, the maximum size of the optimization problem is fixed. Considering the worst case solution time for the optimization is 'k' milliseconds, the total worst case computation time for all agents in the environment is Nk milliseconds on a single core.
\subsection{Comparision with Bounding Volume Expansion}
In Table~\ref{tab:vsKalmanPLR2}, we compare the non-Gaussian formulation with a conservative method based on bounding volume expansion (DCAD~\cite{DCAD}). We observed the bounding volume formulation returned infeasible frequently owing to its conservative approximation. This was observed for the $8$ and $10$ agent cases. A 100 trial runs in the circular scenario was used to tabulate this result. Thus, a conservative method may not be practical in all scenarios. 

\begin{table*}[h]
\centering
%\resizebox{0.8\textwidth}{!}{\begin{minipage}{\textwidth}
    \begin{tabular}{|c|c|c|c|c|c|c|c|c|c|c|}
    \hline
        \multirow{2}{*}{Method} & \multicolumn{5}{c|}{Path Length} & \multicolumn{5}{c|}{No. of trials with collision}\\
        \cline{2-11}
        & 2 & 4 & 6 & 8 & 10 & 2 & 4 & 6 & 8 & 10\\
        \hline
        DCAD with Kalman filter & 41.39 & 45.47 & 61.83 & - & - & 0 & 0 & 4 & - & -\\ 
        non-Gaussian SwarmCCO (n=2) & 41.24 & 45.34 & 47.29 & 51.30 & 57.90 & 0 & 0 & 0 & 6 & 11 \\ 
        \hline
    \end{tabular}
%    \end{minipage}}
    \caption{Comparison of average path length and collision prbability between DCAD (kalman filter) and non-Gaussian SwarmCCO}
    \label{tab:vsKalmanPLR2}
\end{table*}

\section{Conclusion, Limitation, and Future Work}
In this paper, we presented a probabilistic method for decentralized collision avoidance among quadrotors in a swarm. Our method uses a flatness-based linear MPC to handle quadrotor dynamics and accounts for the state uncertainties using a chance constraint formulation. We presented two approaches to model the chance {constraints}; the first assumes a Gaussian distribution for the state, while the second approach is more general and can handle non-Gaussian noise using a GMM. Both the Gaussian and non-Gaussian methods {result in fewer collisions as compared to the deterministic algorithms}, but the Gaussian method was found to be more conservative, leading to longer path lengths for the agents. Further, we {observed} that the non-Gaussian method with 3 Gaussian components {demonstrates better performance in terms of satisfying the ORCA constraints}, resulting in {{fewer}} trials with {an} inter-agent safety distance lower than $1m$ compared to the Gaussian formulation.  %Thus in more constricted environments, non-Gaussian method would perform better since the Gaussian method may end up having an infeasible solution due to conservative behavior. 
On average, our Gaussian method {required} $\sim5ms$ to compute a collision avoiding input, while our non-Gaussian method requires $\sim9ms$ in {scenarios with $4$ agents}.

Our method has a few limitations. { We estimate the distribution of $\mathbf{m}$ using position and velocity samples from a black-box simulator, hence the distribution may not be accurate.} The non-Gaussian method is computationally expensive, {which} affects the rapid re-planning of trajectories. {In addition}, we do not consider the ego-motion noise, i.e. the noise in implementing the control input. Moreover, our optimization's cost function uses mean values and does not consider the uncertainties in the {state}. 

As a part of our future work, we plan to work on faster methods to evaluate the chance constraint for the non-Gaussian, non-parametric case. {Additionally}, we plan to evaluate our algorithm on physical quadrotors.

\bibliographystyle{IEEEtran} 
\bibliography{biblio}

\addtolength{\textheight}{-12cm}   % This command serves to balance the column lengths

\end{document}